\documentclass[letterpaper, 10 pt, conference]{ieeeconf}  

\IEEEoverridecommandlockouts                              

\usepackage{caption}
\usepackage{subcaption}

\usepackage{cite}
\usepackage{colortbl}
\usepackage{xcolor}
\usepackage{graphicx}
\usepackage{wrapfig}
\usepackage[font=footnotesize]{subcaption}
\usepackage{svg}
\usepackage{changepage}
\usepackage{amsmath}
\usepackage{algorithm}
\usepackage{amssymb}
\usepackage{amsmath}
\usepackage{algorithm}
\usepackage[noend]{algpseudocode}
\usepackage{resizegather}
 
\makeatletter
\def\BState{\State\hskip-\ALG@thistlm}
\makeatother

\newlength\myindent
\title{\LARGE \bf
Design and Evaluation of a Hair Combing System \\Using a General-Purpose Robotic Arm
}

\author{Nathaniel Dennler$^{1}$, Eura Shin$^{2}$, Maja Matari\'c$^{1}$ and Stefanos Nikolaidis$^{1}$
\thanks{$^{1}$Nathan Dennler, Maja Matari\'c, and Stefanos Nikolaidis are from the Computer Science Department at the University of Southern California, Los Angeles, CA
        {\tt\small \{dennler, mataric, nikolaid\} @usc.edu}}%
        
\thanks{$^{2}$Eura Shin is from the Computer Science Department at Harvard University, Cambridge, MA
        {\tt\small eurashin@g.harvard.edu}}%
}

\begin{document}








\maketitle
\thispagestyle{empty}
\pagestyle{empty}

\begin{abstract}
    This work introduces an approach for automatic hair combing by a lightweight robot. For people living with limited mobility, dexterity, or chronic fatigue, combing hair is often a difficult task that negatively impacts personal routines. We propose a modular system for enabling general robot manipulators to assist with a hair-combing task. The system consists of three main components.  The first component is the segmentation module, which segments the location of hair in space. The second component is the path planning module that proposes automatically-generated paths through hair based on user input. The final component creates a trajectory for the robot to execute. We quantitatively evaluate the effectiveness of the paths planned by the system with 48 users and qualitatively evaluate the system with 30 users watching videos of the robot performing a hair-combing task in the physical world. The system is shown to effectively comb different hairstyles.
    
    
\end{abstract}
\section{Introduction}


There are 40.7 million people in United States living with some form of reduced mobility \cite{nhis_2019} resulting from a variety of factors (e.g., traumatic injury, stroke, genetic causes, etc.), according to the National Health Interview Survey. Most items used in activities of daily living (ADLs) are not explicitly designed for people living with reduced mobility and may be completely unusable. This requires people living with limited mobility to either purchase specialized items designed for mobility limitations or acquire assistance from trained professionals.  However, for many people, neither option may be accessible nor affordable.

The high costs of formal caregivers has necessitated their replacement with family members who give care informally. In 2015, 43.5 million people in the US reported providing informal care in the previous 12 months. Informal caregiving requires providing companionship, personal care, scheduling health services, providing transportation, and more, while at the same time managing one's own job, family finances, household tasks, and other daily activities \cite{feinberg2011valuing}. Due to the increased caregiver load, informal caregiving can be a source of stress and can lead to health problems \cite{pinquart2003differences}. However, assisting people with limited mobility with ADLs is necessary to decrease their risk of depression and, in some forms of limited mobility, can aid in their rehabilitation processes \cite{lin2011does}.


Assistive robots could provide a solution to this growing issue by enabling people with limited mobility to independently perform necessary daily tasks such as eating \cite{bhattacharjee2020more}, drinking \cite{goldau2019autonomous}, and grabbing objects \cite{jain2010assistive}. Areas of self-care such as shaving \cite{hawkins2014assistive}, cleaning \cite{king2010towards}, and dressing \cite{erickson2018deep,kapusta2019personalized} have similarly shown potential for assistive robotics for non-critical tasks. However, fully autonomous systems are not always desirable, as it limits the user's choice and leads to lower acceptance of the system in the event of errors \cite{bhattacharjee2020more}. A key design goal is thus to include user input to the system in a way that reduces the effort of the user yet simultaneously maintains the user's own autonomy.



We propose to expand the abilities of general-purpose assistive robotic arms to the domain of hair-combing, an important and under-explored self-care ADL. In this work, we provide 
a minimal system to comb different kinds of hairstyles. The system consists of three modules: an image segmentation module, a path-planning module, and a trajectory generation module.  The main contributions of this paper are: 1) the formulation of a path planning method for hair combing, and 2) insights from na\"ive users on the efficacy of that approach to hair combing. Together, the system provides users or remote caretakers a way to automatically generate paths through a user's hair and have a robot comb along the generated paths. The system provides functionality using a
robotic arm and an RGB-D camera. This work presents a physical implementation of the system on a lightweight Kinova Gen2 robot arm and an evaluation of its performance. Our results demonstrate that the system successfully combs a variety of hairstyles. 


\section{Related Work}

\begin{figure}[t]
  \centering
\includegraphics[width=.9\linewidth]{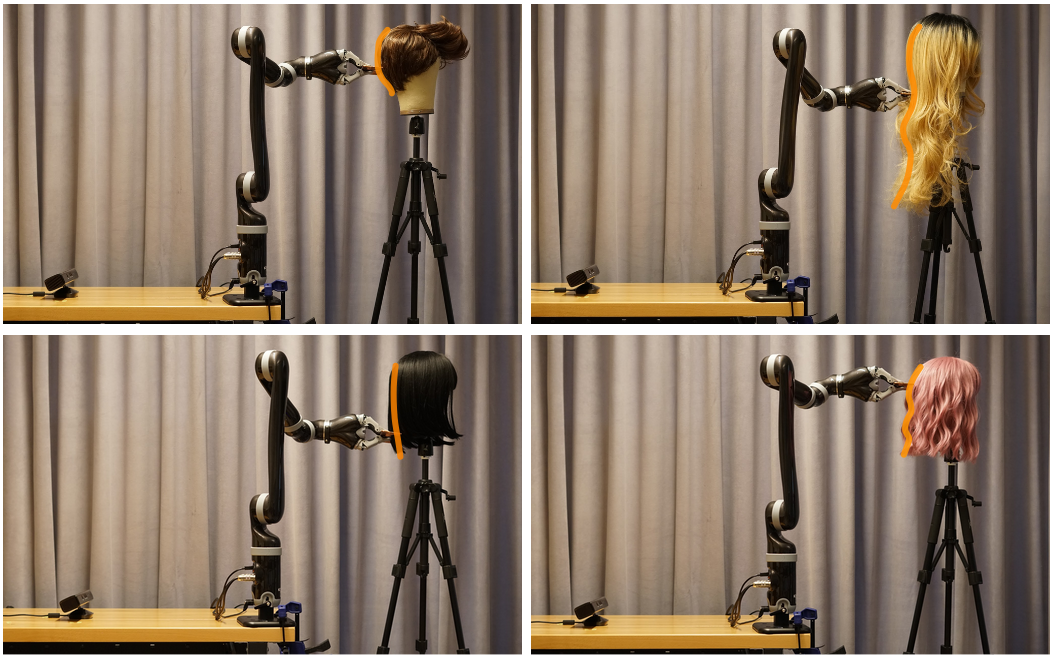}
  \caption{System setup and trajectories for different hairstyles.}
   \label{fig:setup}
\end{figure}



\textit{Assistive Robotics}:
Advances have made it possible to provide robot-assisted ADLs \cite{brose2010role}. Many of the existing systems are high degree-of-freedom arms mounted on tables~\cite{candeias2018vision,jardon2012functional}, wheelchairs~\cite{alqasemi2005wheelchair,schrock2009design,gallenberger2019transfer}, or mobile platforms \cite{park2016towards,rhodes2018robot} capable of accomplishing a wide variety of tasks. Several are commercially available \cite{kinova, performance, assistive}.
However, these systems are often difficult to control due to the high number of controllable joints, creating a need for some autonomy in order to perform ADLs in an efficient and low-effort manner \cite{herlant2016assistive}. In addition, many of these systems require specialized or wearable sensors that can have unintended negative physical and psychological consequences for the user \cite{schukat2016unintended}. To mitigate those issues, our work aims to be minimally invasive and uses a minimal setup, while additionally utilizing a general-purpose robot arm that is useful in other capacities. 

\textit{Hair Modeling}:
Current works in hair modeling focus on recreating and rendering entire hair structures in 3D from 2D images \cite{ward2007survey}. 
In general, hair can be modeled as a collection of curves in 3D space that follow an underlying orientation field that can be inferred from the surface of the hair \cite{saito20183d, zhang2019hair}. Some approaches utilize a database of 3D models, which can be combined to form new styles that are found in single-view images \cite{hu2015single}. Other approaches \cite{chai2012single} use a pipeline for editing hairstyles in portraits by finding orientation fields from images to generate additional strands \cite{chai2012single}. Our system takes inspiration from these works, and models hair as an orientation field. We emphasize being able to generate paths in real time on hair that becomes neater as the system combs it. 

\section{System Description}

Our goal is to design a system for people living with limited mobility so they can comb their own hair. 
To do this, we propose a system consisting of three modules: the \textit{segmentation module} localizes the hair in space, the \textit{path planning} module proposes brush strokes from a starting point input by the user, and the \textit{trajectory generation module} provides the robot with a trajectory through space to move along the user-selected path.

\subsection{Hardware}
The hardware for this system consists of a depth camera and a robot manipulator. Our implementation uses a Microsoft Kinect v1 depth sensor which is commonly used in various robotics domains.  We used the Kinova JACO Gen 2 lightweight robotic manipulator, which is also used effectively in several other assistive settings \cite{maheu2011evaluation, herlant2016assistive} because it is safe for physical interaction. While a comb is also required, our system is designed to work with any comb that can be grasped by the robot arm. We chose a wide-toothed comb, but the comb should be chosen with consideration for the user's hair type. A diagram of the system setup for different hairstyles is shown in Fig. \ref{fig:setup}.


\subsection{Segmentation Module}

\begin{figure}[t!]
  \centering
  \includegraphics[width=.9\linewidth]{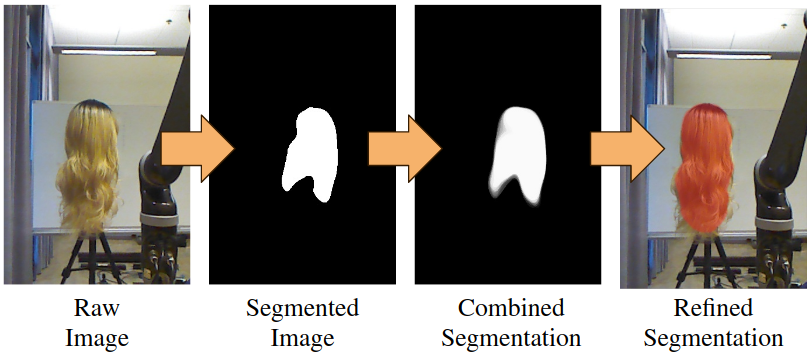}
  \caption{Diagram of the image segmentation module.}
   \label{fig:segmentation}
\end{figure}

Hair is difficult to locate in images due to its high variability in color, length, and style. We use a learning-based approach to segment hair in RGB images and consider this mask a noisy estimate of the hair. We refine this estimate through signal processing techniques as outlined in Fig. \ref{fig:segmentation}

\subsubsection{Segmentation Formulation}

Segmentation is a well-established field of study in computer vision \cite{haralick1985image,ghosh2019understanding}. The overall approach for neural architectures is to take a high dimensional image and learn a low-dimensional representation of the image with an encoder network. The low-dimensional representation is then up-sampled to form an image of the same size as the original image that contains the mask. The loss is quantified as the binary cross-entropy loss between a ground-truth mask and the predicted mask. 


Due to the success and efficiency of fine-tuning pre-trained models, we use SqueezeNet \cite{iandola2016squeezenet} as the encoder; it is a small-sized network that achieves high accuracy on Image Net Classification tasks. The smaller model provides comparable accuracy with fewer parameters and thus has a faster run time, which is essential for real-time applications like ours. The decoder has a similar architecture to the encoder, but uses deconvolutional layers instead. We used the Figaro 1k dataset \cite{umar2018hair} for training; it contains a wide variety of labelled hair segments from unconstrained camera angles.

\subsubsection{Refining the segmented region}

To improve the fidelity of the mask, we extract the points in XYZ space that correspond with the hair pixels. Two main sources of error are of concern in accurately segmenting hair from RGB images: \textit{temporal} errors and \textit{spatial} errors. 

Temporal errors can result from sensor noise, causing the estimated hair mask to form discontinuities or disappear completely for a single frame. 
To reduce the impact of such high-frequency noise, we use an exponential smoothing filter to average values from the stream of masks. This filter averages masks over time, but gives exponentially higher weights to images nearby in the sequence. This reduces the effects of high-frequency noise (i.e., sensor noise) but allows low-frequency patterns (i.e., user movement) to pass through. 
The filter is described by the following iterative function:

\begin{equation}
mask(t, \hat{Y}_t) = \left\{
        \begin{array}{ll}
            \hat{Y}_t & \quad t = 0 \\
            \alpha \cdot \hat{Y}_{t} + (1-\alpha) \cdot mask(t-1, \hat{Y}_{t-1}) & \quad t > 0
        \end{array}
        \right\}
\end{equation}

where $t$ is the timestep, $\hat{Y}_t$ is the output of the segmentation network for frame , and $\alpha$ is the smoothing parameter. Higher values of $\alpha$ allow better tracking of user movement and lower values reduce the effect of sensor noise. Our system uses $\alpha=0.9$.

To address spatial errors (i.e., false-positives, other people in the frame, etc.) we first select the largest contiguous area in the mask. We then find the median value of pixels selected as representing the depth of the hair,  an estimate that is robust to outliers. Points that are within two standard deviations of this median value and have hues within the range of the pixels in the mask are added to the mask to reduce the number of false negatives.



\subsection{Path Planning Module}

We developed a path planning module that takes the current hair state from the segmentation module and a user-specified point as input and produces a path through the hair that aligns with the natural curvature of the hair. For this work, we use a mouse to make the selection, but this pipeline is not dependent on the input device. Following works from Computer Graphics on hair rendering, we model the flow of hair as a differential field \cite{chai2012single}. To obtain the differential field, we find the orientation of the hair in the image.
We consider a path that follows the flow of the hair as a solution to the differential field, with initial conditions defined as the point a user selects as the path's starting point.
For the following sections, we leverage the common assumption that an image is a smooth continuous function that maps XY coordinates to intensity values. The function is discretized into pixels and the intensity value at each pixel is noisily sampled from the underlying intensity function. This assumption allows us to take derivatives over the image to construct the orientation field.


\subsubsection{Coherence-Enhancing Shock Filters} 

We first use the method described by Weickert et al. \cite{weickert2003coherence} to make the natural flows of the hair more coherent, reducing the effect that stray strands of hair on the calculation of orientation. To do this, we iteratively refine an image by increasing contrast in the direction of the greatest change of the intensity gradient (directions perpendicular to the flow of hair), and decreasing contrast in the direction of the lowest change of the intensity gradient. 

\begin{figure}[t!]
  \centering
  \includegraphics[width=0.8\linewidth]{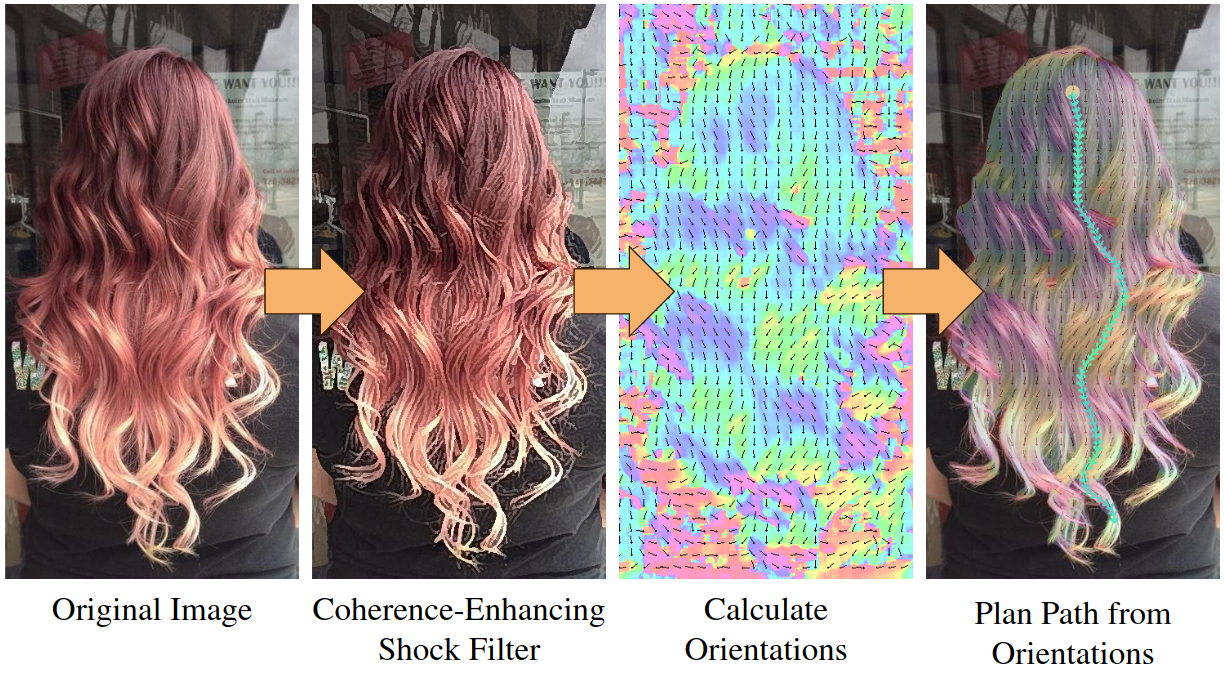}
  \caption{Overview of path-planning module.}
   \label{pathplanningpipeline}
\end{figure}

First, we calculate the dominant eigenvector $e$ for each pixel's local neighborhood, which provide the direction of greatest change in intensity. 
The convexity in the direction of $e$ is determined by $e^THe$, where $H$ is the Hessian matrix in each pixel's neighborhood.
If the convexity is positive, the algorithm intensifies the pixel to match its surroundings. If the convexity is negative, the algorithm reduces the pixels intensity. The net effect of this process is to increase the contrast in the direction of the greatest change in intensity. The process is shown in Algorithm \ref{shockFilter}, and the effect on images is seen in Fig. \ref{pathplanningpipeline}.




\begin{algorithm}
\caption{Coherence Filter}\label{shockFilter}
\begin{algorithmic}[1]
\State \textbf{Input:} Image $I$, kernel size $K_\delta$ for approximating derivative, kernel size $K_e$ for calculating eigenvectors, kernel size $K_m$ for calculating max and min values, Constant $C_{blend}$ the blending rate for each iteration, and $T$ the total number of iterations.

\For{$t=1,2,..T$}
        \State Calculate first (normalized) eigenvector $[e_x,e_y]$ for each sub image of size $K_e \times K_e$ in $I_t$
        \State Approximate $\frac{\delta^2 I}{\delta x^2}$, $\frac{\delta^2 I}{\delta x\delta y}$, and $\frac{\delta^2 I}{\delta y^2}$ with a sobel filter of size $K_\delta$.
        \State Compute  $I_{vv} \leftarrow e_x^2*\frac{\delta^2 I}{\delta x^2} +2*e_x*e_y*\frac{\delta^2 I}{\delta x\delta y} + e_y^2*\frac{\delta^2 I}{\delta y^2}$
        \State Create a new image $I_t'$, where for each sub image  $K_m \times K_m$ in $I$, we take the max value of the sub image if $I_{vv} > 0$ and the min if $I_{vv} < 0$ 
        \State $I_{t+1} \gets I_t*C_{blend} + I_t'*(1-C_{blend})$
      \EndFor
\State \textbf{return} \text{image}
\State \textit{Note: our implementation uses $K_{\delta}=7$, $K_{e}=11$, \Statex  $K_m=3$, $C_{blend}=0.9$, and $T=3$}
\end{algorithmic}
\end{algorithm}

\begin{algorithm}
\caption{Find Orientations}\label{GST}
\begin{algorithmic}[1]
\State \textbf{Input:} Image $I$, kernel size $K_\delta$ for approximating derivative, kernel size $K_E$ for approximating expectation

\State Approximate $\frac{\delta I}{\delta x}$ and $\frac{\delta I}{\delta y}$ (with a sobel filter of size $K_\delta$).
\State Construct:
$\begin{bmatrix}
I_xI_x & I_xI_y \\
I_yI_x & I_yI_y 
\end{bmatrix} = \nabla I^T \nabla I = S_0$ 
\State Locally average each element in $S_0$ with kernel size $K_E$ to calculate Structure Tensor: $\begin{bmatrix}
J_{11} & J_{12} \\
J_{21} & J_{22} 
\end{bmatrix} = S_{K_E}$

\State Calculate local orientation: $\Theta = \frac{1}{2} atan2(\frac{J_{12} + J{21}}{J_{22} - J{11}}) + \frac{\pi}{2}$



\State \textbf{return} $\Theta$
\State \textit{Note: our implementation uses $K_{\delta}=3$ and $K_{E}=5$}
\end{algorithmic}
\end{algorithm}

\begin{algorithm}
\caption{Calculate Path}\label{Path}
\begin{algorithmic}[1]
\State \textbf{Input:} Orientation image $\Theta$, selected pixel for the start of the path $p_0 = [p_x,p_y]$, step size $k$ (in pixels).
\State initialize $path$ with $v_0$
\While {$p_t$ is within bounds of hair}
\State $\theta_{p_t} = \Theta[p_t]$
\State $p_{t+1} \gets p_t + k \cdot [cos(\theta_{p_t}), sin(\theta_{p_t})]$
\State append $p_{t+1}$ to the path
\EndWhile

\State \textbf{return} $path$

\State \textit{Note: our implementation uses $k=6$}
\end{algorithmic}
\end{algorithm}

\subsubsection{Calculating Orientation}
Once the image has been filtered to enhance the coherence of the hair flows, the system calculates the direction of the hair flows in the image. The gradient structure tensor provides this information, which is defined continuously for a point $p$ as:

\begin{gather}
   S_{K_E} = \begin{bmatrix}
    \int_r {K_E}(r) \cdot (I_x(p-r))^2 & \int_r {K_E}(r) \cdot I_x(p-r)I_y(p-r) \\
    \int_r {K_E}(r) \cdot I_y(p-r)I_x(p-r)& \int_r {K_E}(r) \cdot (I_y(p-r))^2 
    \end{bmatrix}
\end{gather}

where $K_E$ represents a window of the part of the image to consider. The integral corresponds to the convolution operation over a continuous image. The pseudo-code for computing $S_{K_E}$ discretely for all pixels is detailed in Algorithm \ref{GST} on lines 2-4. Orientations from this structure tensor following the equation on line 5, as detailed by Yang \cite{yang1996structure}.

\subsubsection{Generating the path}

Once the orientation field is constructed, the user selects a point on the image for the starting point for the comb. Once the point is selected, the path is iteratively generated by adding new points that follow the direction of the orientation field, as described in Algorithm \ref{Path}.

\subsection{Trajectory Generation Module}

The trajectory generation module takes in the path in the form of pixel coordinates and maps them to task-space poses.
While we used a stationary camera for the mapping, this is not required, since we only need the transformation between the camera and robot frames when the image was taken. 

We retrieve the XYZ position of each point in the path from the point cloud, where we index points in the point cloud by the pixels in the path.  We assign  orientations to each point, so that the teeth of the comb (z-axis of the end-effector, following the Denavit–Hartenberg convention) point orthogonally to the user's hair plane $H$,  formed by the visible portion of the hair. Its normal vector is $\vec{h}$.

\begin{figure}[t!]
  \centering
  \includegraphics[width=0.85\linewidth]{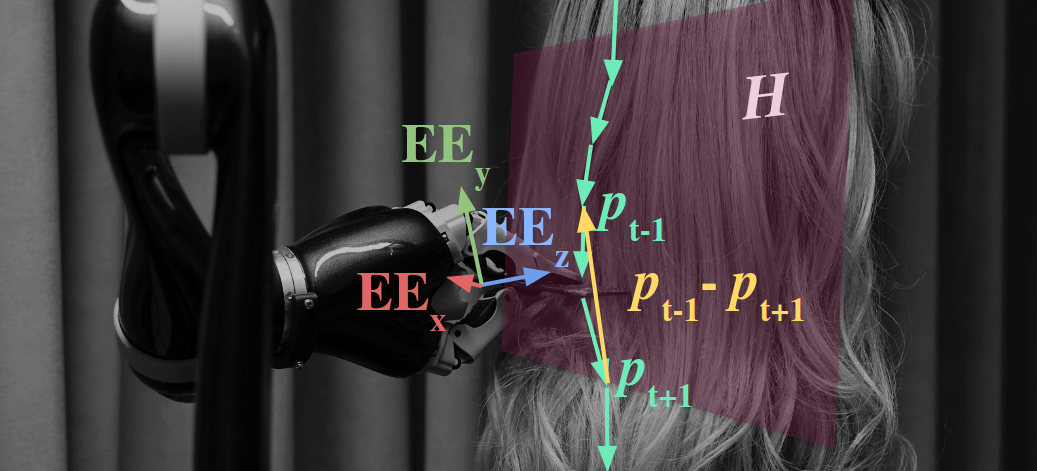}
  \caption{Creating task-space poses from points of stroke.}
   \label{fig:geometrySetup}
   \vspace{-0.2cm}
\end{figure}

The Y axis of the end effector is aligned with the vector tangent to the path that the robot is following, projected onto the hair plane. This is approximated with the centered difference of the previous and next points in the path. Figure~\ref{fig:geometrySetup} shows this graphically, where the direction of the end effector's y coordinate frame before normalization (measured from the camera's reference frame) at point $p_t$ is:

\begin{gather}
EE_y = (\vec{p}_{t-1} - \vec{p}_{t+1}) - \frac{(\vec{p}_{t-1} - \vec{p}_{t+1}) \cdot \vec{h}}{\Vert\vec{h}\Vert}
\end{gather}

A motion plan for the robot end effector is planned in task space, following joint velocity and acceleration constraints. A preview of the resulting motion plan is shown to the user. If the user accepts the preview, the robot executes the motion plan, otherwise the system prompts the user to select a new start point in the hair and generates a new path and trajectory. 

\begin{figure}[t!]
  \centering

 \begin{subfigure}{0.45\linewidth}
      \includegraphics[width=\linewidth]{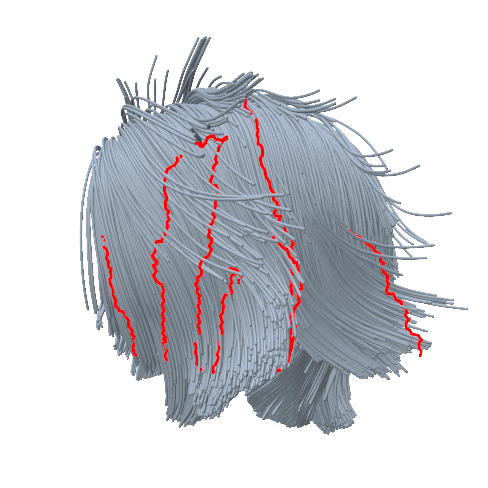}
      \caption{Mesh-based Algorithm.}
    \end{subfigure}\hfil
    \begin{subfigure}{0.45\linewidth}
      \includegraphics[width=\linewidth]{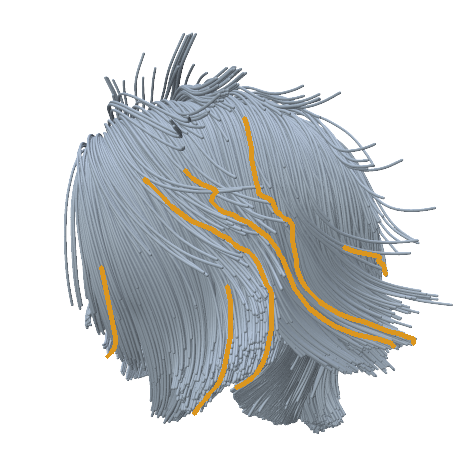}
       \caption{Image-based Algorithm.}
     \end{subfigure}\hfil
    
  \caption{Illustrative hairstyle where the image-based algorithm performs differently than the mesh-based for corresponding starting points.}
   \label{fig:meshPaths}
\end{figure}

\begin{figure*}[t]
    \centering
    
    \begin{subfigure}{0.095\textwidth}
      \includegraphics[width=\linewidth]{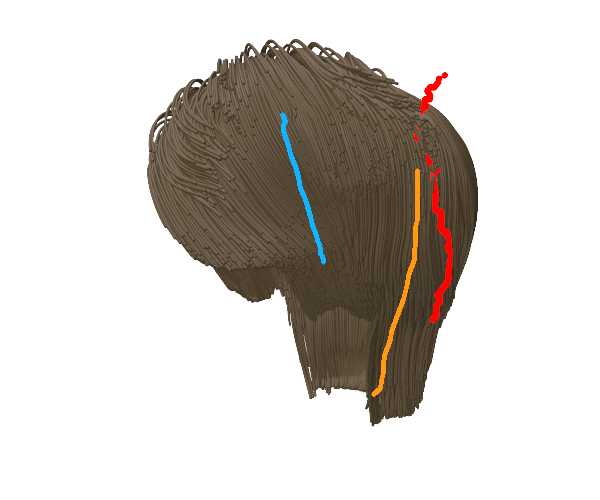}
    
    \end{subfigure}\hfil 
    \begin{subfigure}{0.095\textwidth}
      \includegraphics[width=\linewidth]{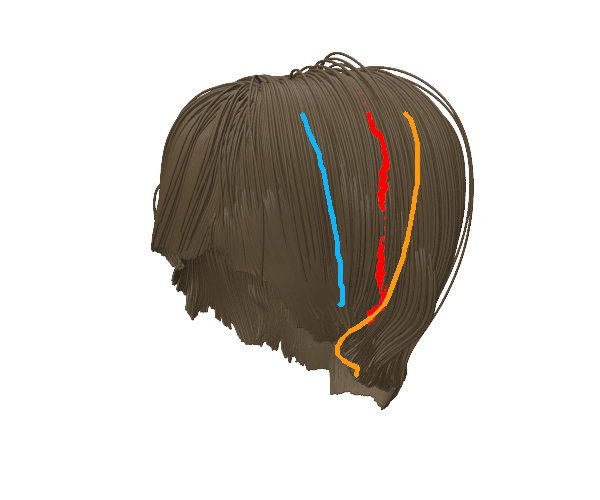}
    
    \end{subfigure}\hfil 
    \begin{subfigure}{0.095\textwidth}
      \includegraphics[width=\linewidth]{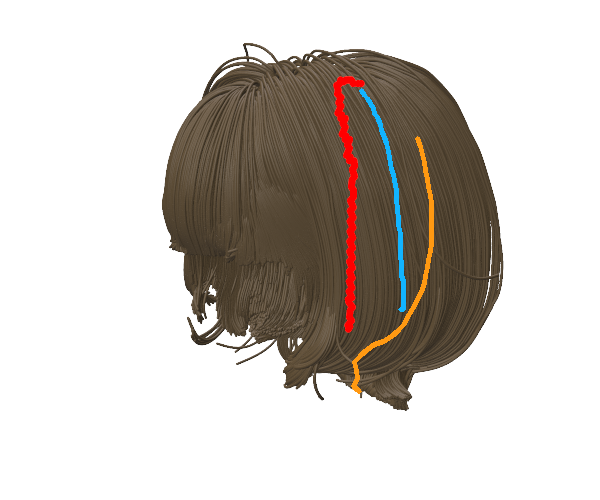}
    
    \end{subfigure}\hfil 
    \begin{subfigure}{0.095\textwidth}
      \includegraphics[width=\linewidth]{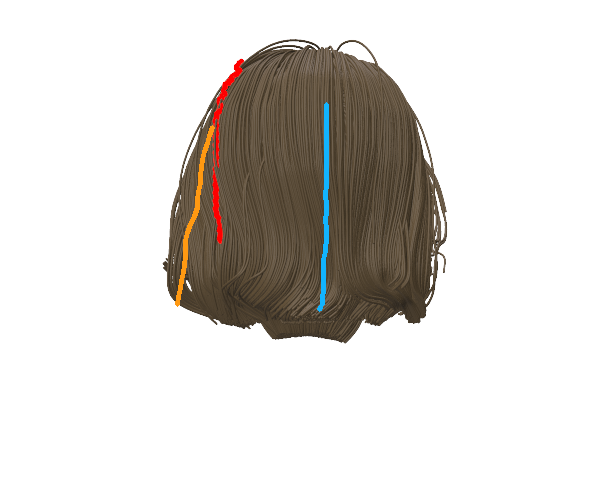}
    
    \end{subfigure}\hfil 
    \begin{subfigure}{0.095\textwidth}
      \includegraphics[width=\linewidth]{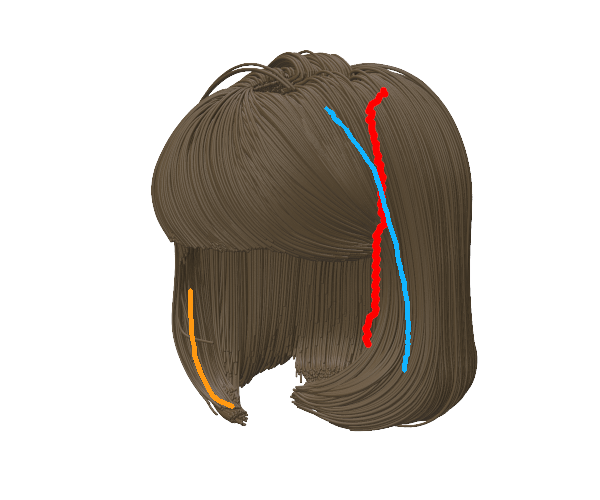}
    
    \end{subfigure}\hfil 
    \begin{subfigure}{0.095\textwidth}
      \includegraphics[width=\linewidth]{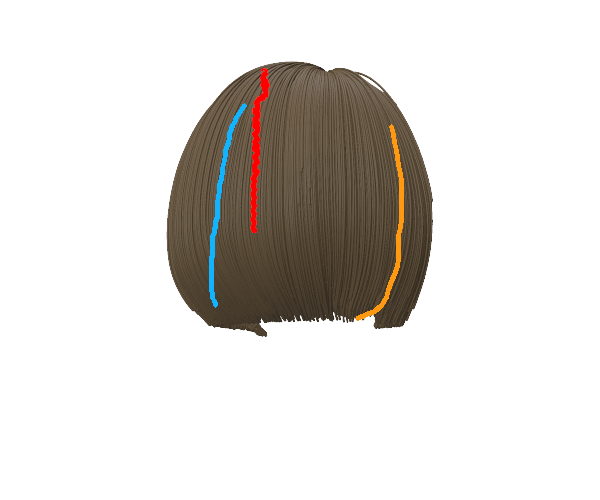}
    
    \end{subfigure}\hfil 
    \begin{subfigure}{0.095\textwidth}
      \includegraphics[width=\linewidth]{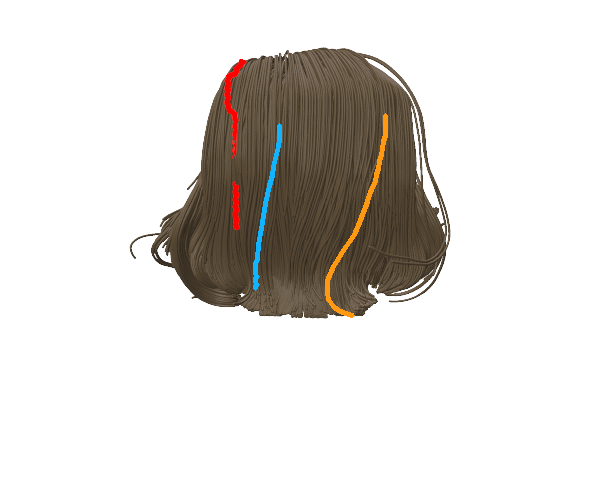}
    
    \end{subfigure}\hfil 
    \begin{subfigure}{0.095\textwidth}
      \includegraphics[width=\linewidth]{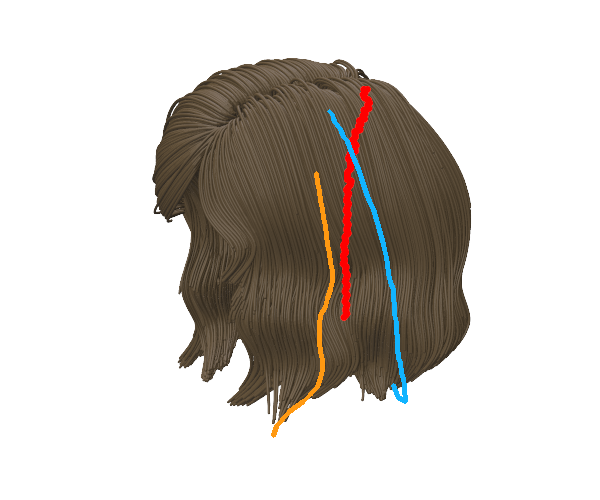}
    
    \end{subfigure}\hfil 
    \begin{subfigure}{0.095\textwidth}
      \includegraphics[width=\linewidth]{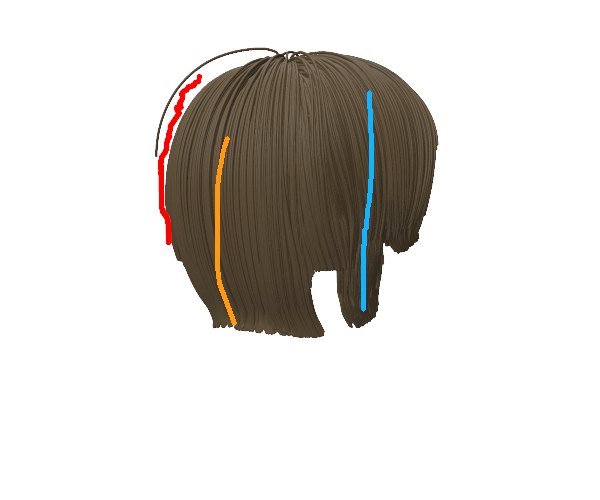}
    
    \end{subfigure}\hfil 
    \begin{subfigure}{0.095\textwidth}
      \includegraphics[width=\linewidth]{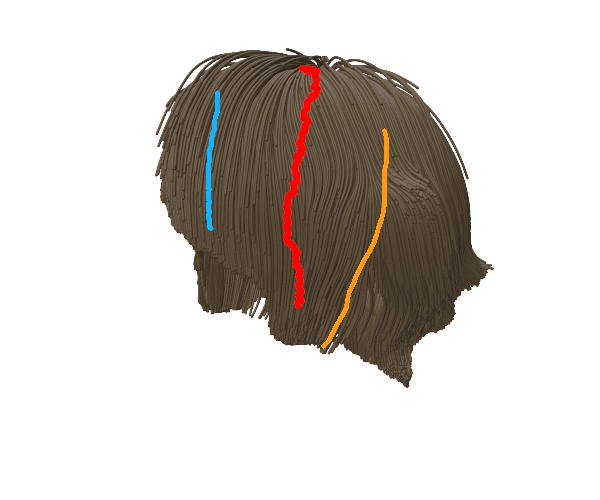}
    
    \end{subfigure}\hfil 
    
    \medskip
    
    \begin{subfigure}{0.095\textwidth}
      \includegraphics[width=\linewidth]{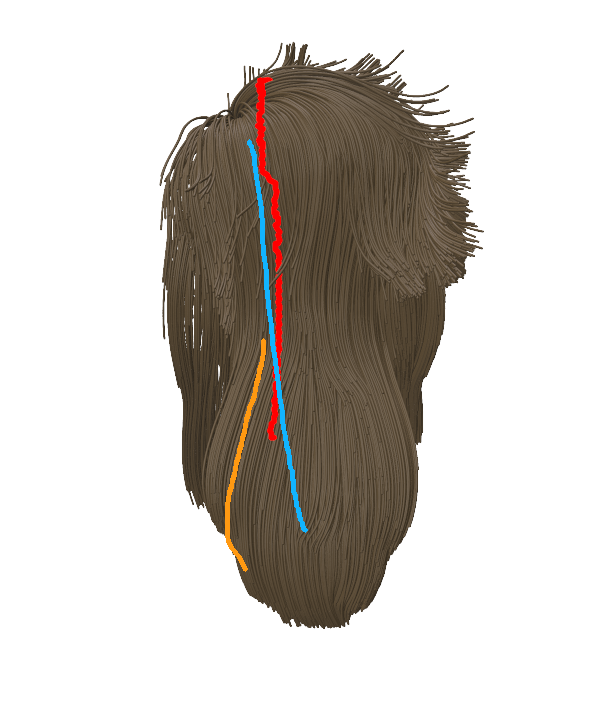}
    
    \end{subfigure}\hfil 
    \begin{subfigure}{0.095\textwidth}
      \includegraphics[width=\linewidth]{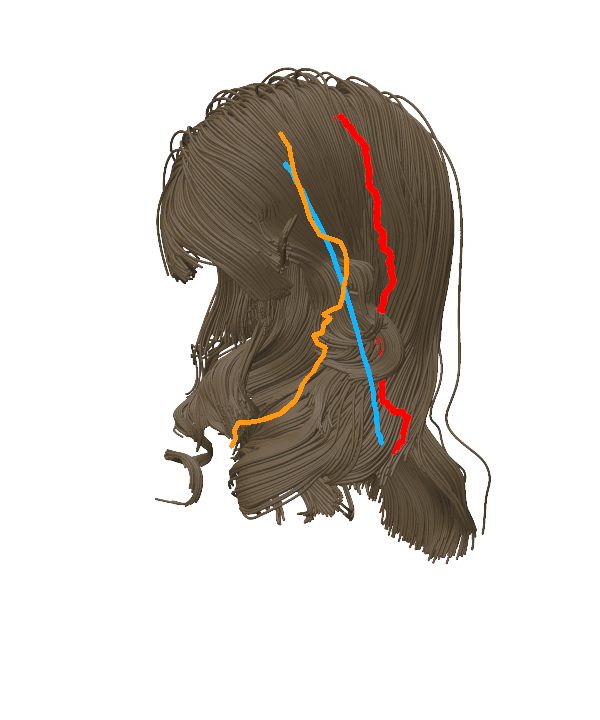}
    
    \end{subfigure}\hfil 
    \begin{subfigure}{0.095\textwidth}
      \includegraphics[width=\linewidth]{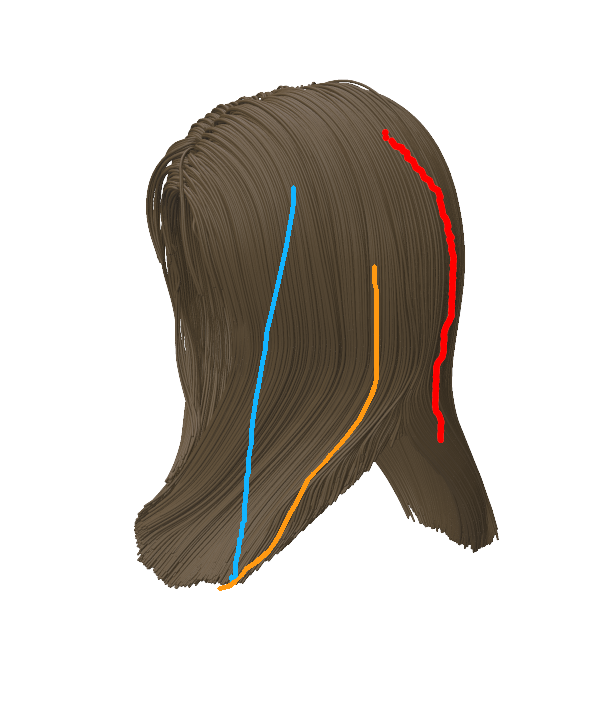}
    
    \end{subfigure}\hfil 
    \begin{subfigure}{0.095\textwidth}
      \includegraphics[width=\linewidth]{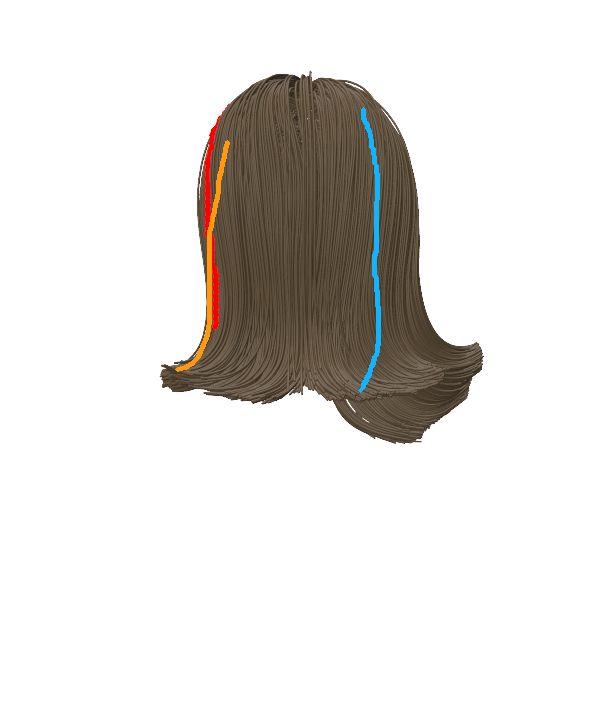}
    
    \end{subfigure}\hfil 
    \begin{subfigure}{0.095\textwidth}
      \includegraphics[width=\linewidth]{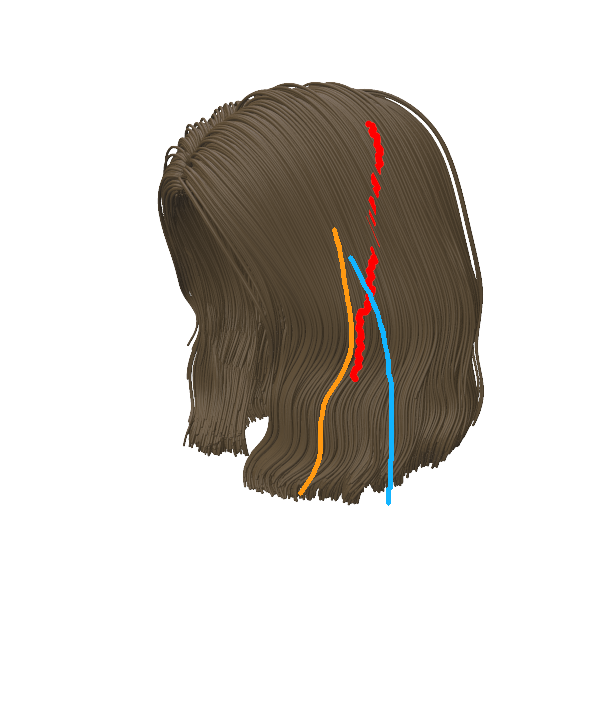}
    
    \end{subfigure}\hfil 
    \begin{subfigure}{0.095\textwidth}
      \includegraphics[width=\linewidth]{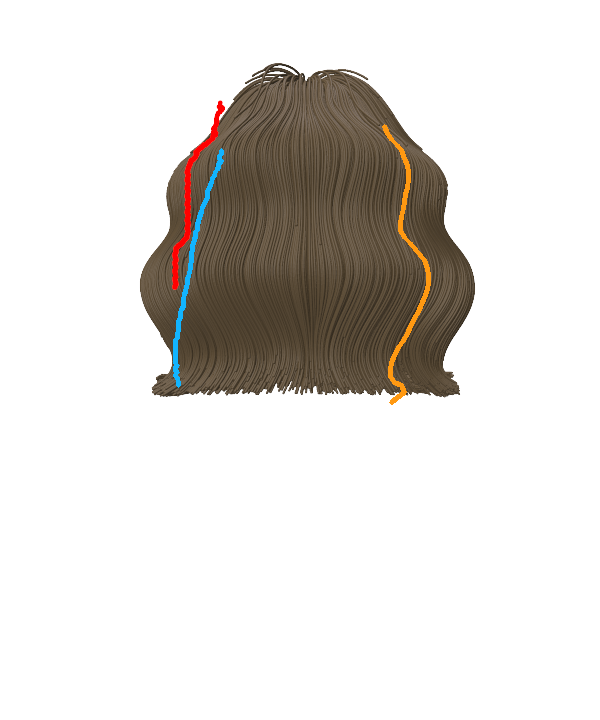}
    
    \end{subfigure}\hfil 
    \begin{subfigure}{0.095\textwidth}
      \includegraphics[width=\linewidth]{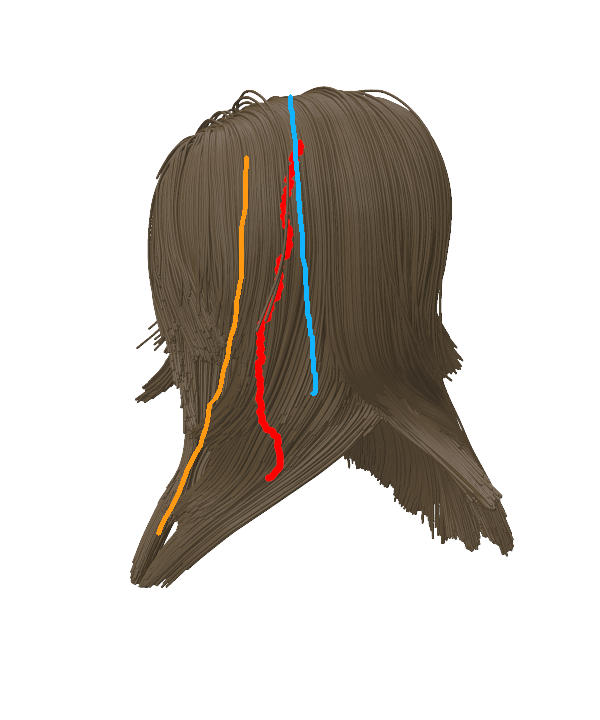}
    
    \end{subfigure}\hfil 
    \begin{subfigure}{0.095\textwidth}
      \includegraphics[width=\linewidth]{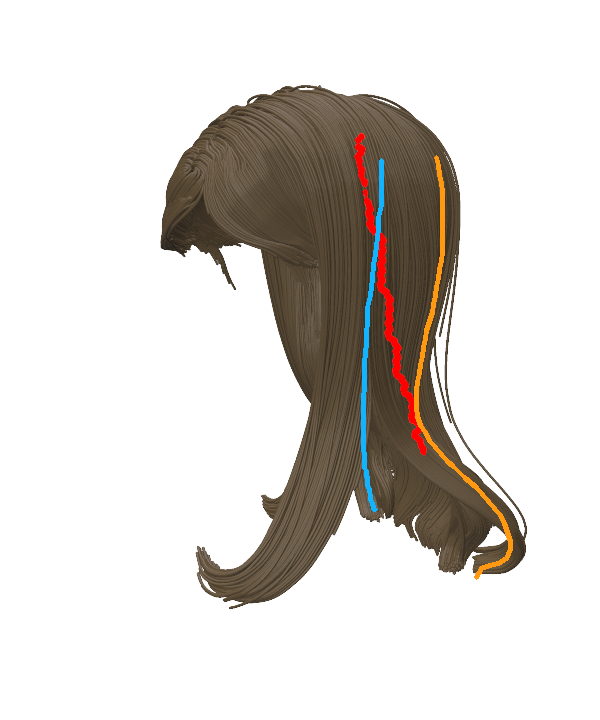}
    
    \end{subfigure}\hfil 
    \begin{subfigure}{0.095\textwidth}
      \includegraphics[width=\linewidth]{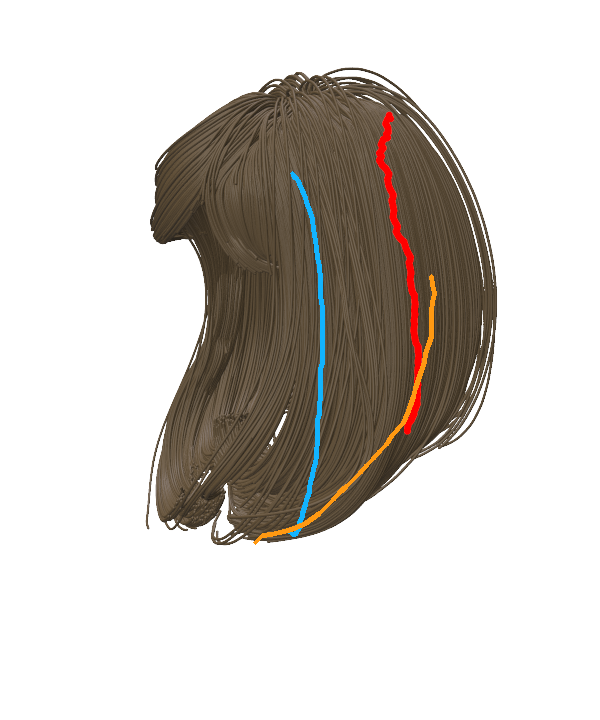}
    
    \end{subfigure}\hfil 
    \begin{subfigure}{0.095\textwidth}
      \includegraphics[width=\linewidth]{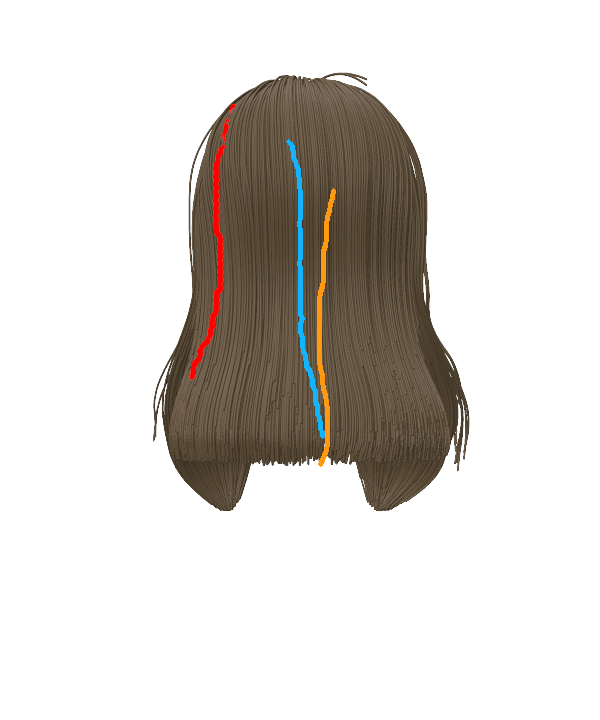}
    
    \end{subfigure}\hfil 

    \medskip
    
    \begin{subfigure}{0.095\textwidth}
      \includegraphics[width=\linewidth]{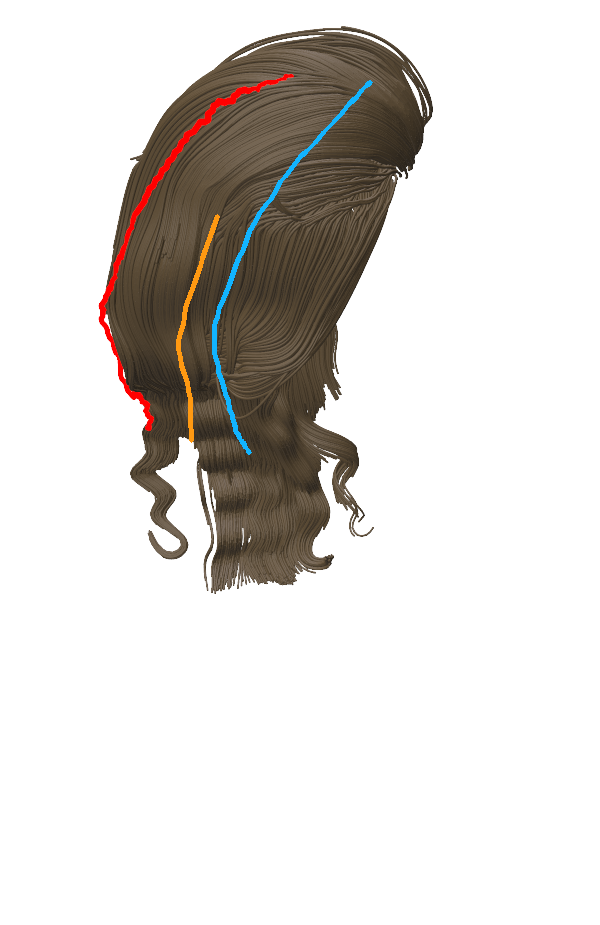}
    
    \end{subfigure}\hfil 
    \begin{subfigure}{0.095\textwidth}
      \includegraphics[width=\linewidth]{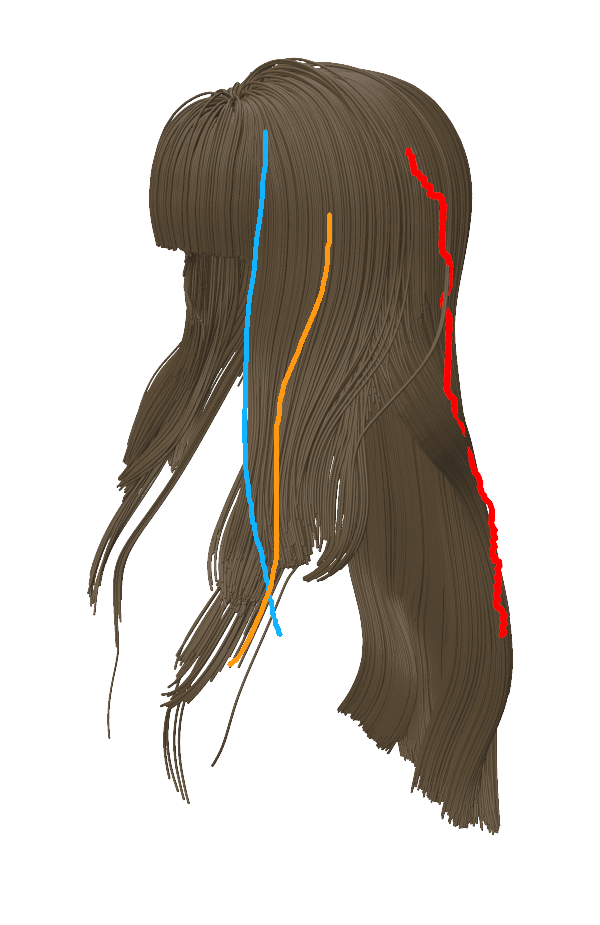}
    
    \end{subfigure}\hfil 
    \begin{subfigure}{0.095\textwidth}
      \includegraphics[width=\linewidth]{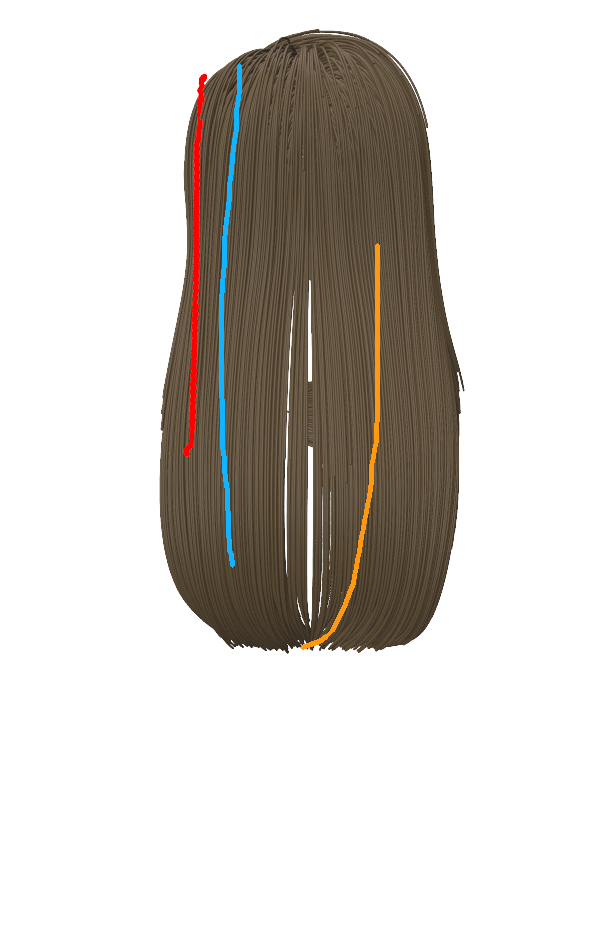}
    
    \end{subfigure}\hfil 
    \begin{subfigure}{0.095\textwidth}
      \includegraphics[width=\linewidth]{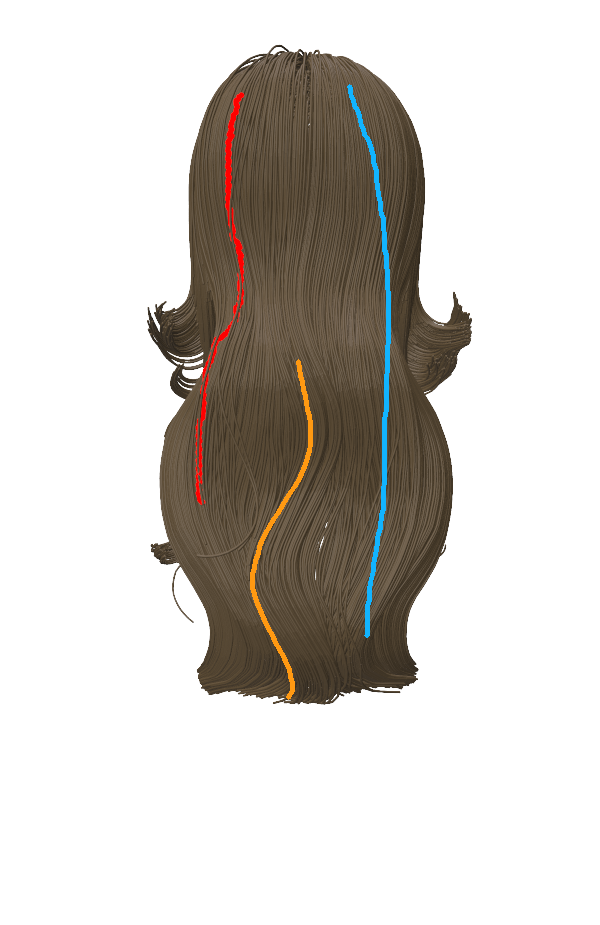}
    
    \end{subfigure}\hfil 
    \begin{subfigure}{0.095\textwidth}
      \includegraphics[width=\linewidth]{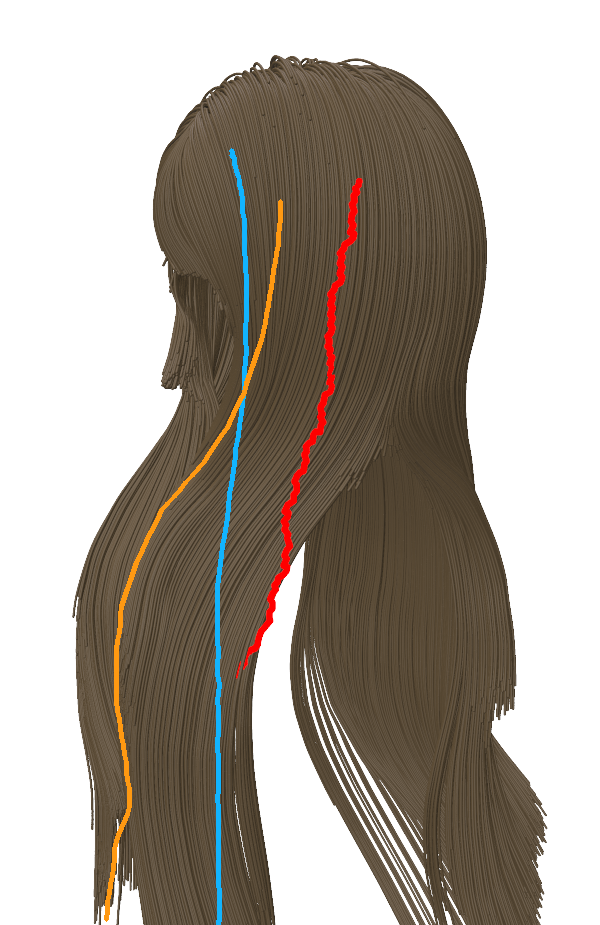}
    
    \end{subfigure}\hfil 
    \begin{subfigure}{0.095\textwidth}
      \includegraphics[width=\linewidth]{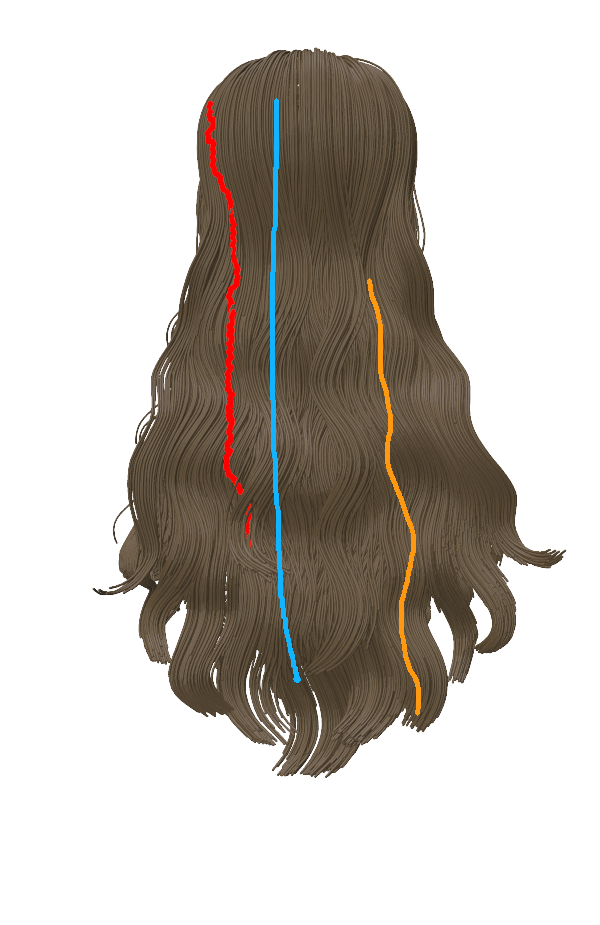}
    
    \end{subfigure}\hfil 
    \begin{subfigure}{0.095\textwidth}
      \includegraphics[width=\linewidth]{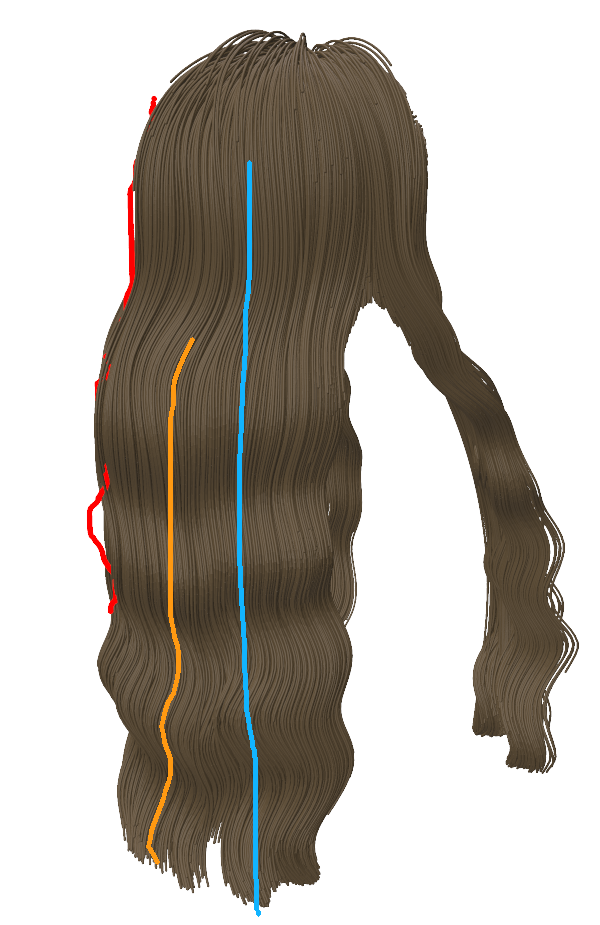}
    
    \end{subfigure}\hfil 
    \begin{subfigure}{0.095\textwidth}
      \includegraphics[width=\linewidth]{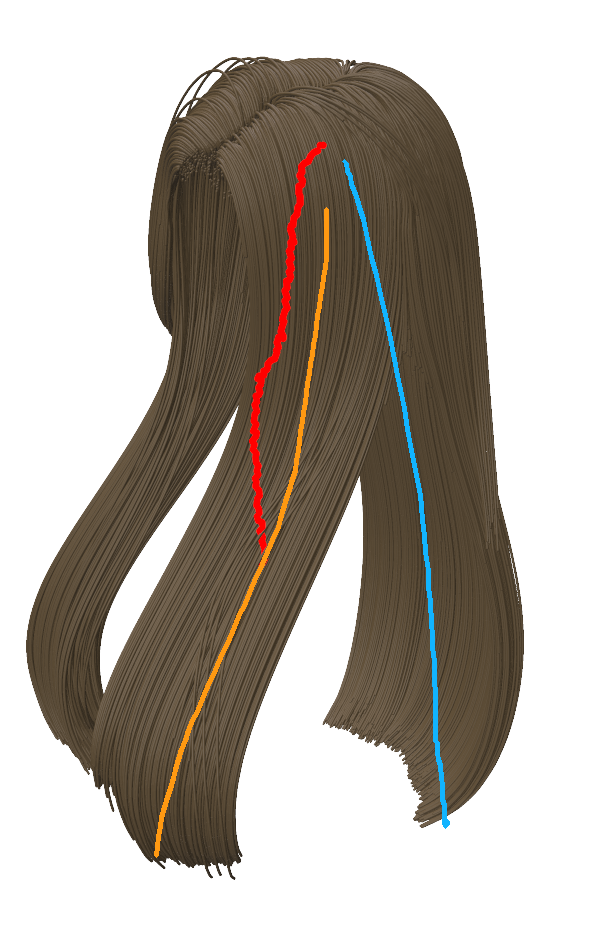}
    
    \end{subfigure}\hfil 
    \begin{subfigure}{0.095\textwidth}
      \includegraphics[width=\linewidth]{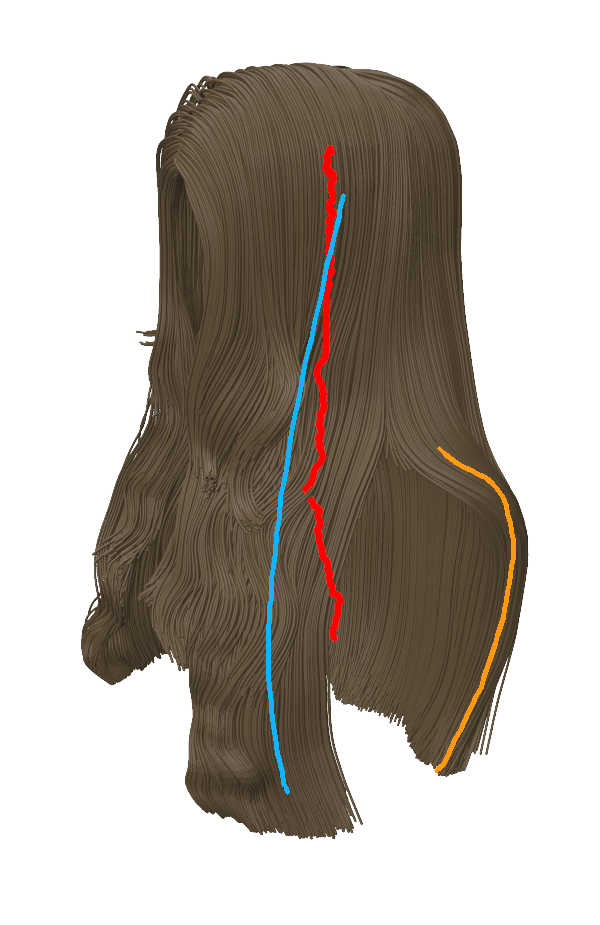}
    
    \end{subfigure}\hfil 
    \begin{subfigure}{0.095\textwidth}
      \includegraphics[width=\linewidth]{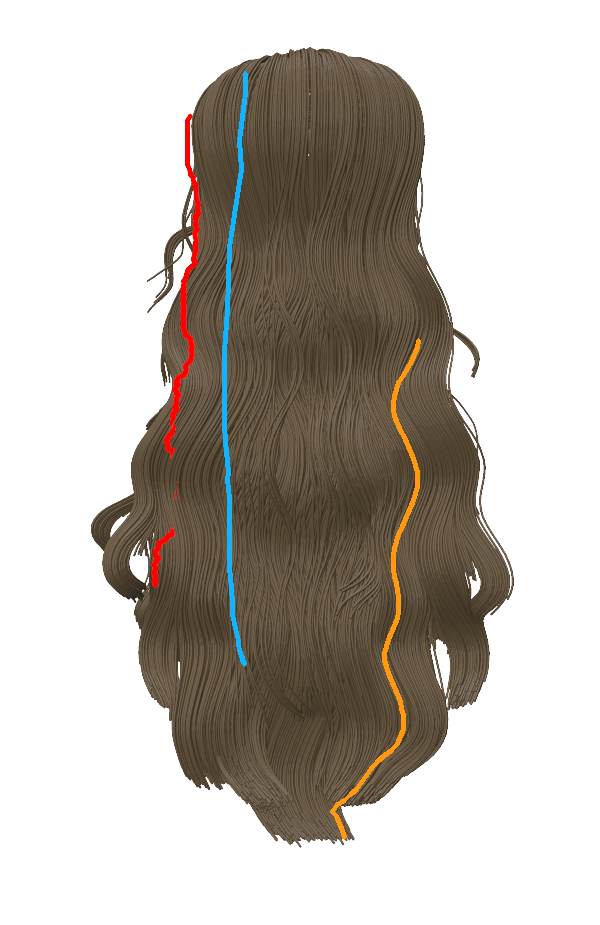}
    \end{subfigure}\hfil 
    \label{hairstyles}
    \caption{Examples of generated paths randomly sampled from the set of solutions on the thirty hairstyles. Each hairstyle came in five different colors and from one of three camera viewpoints used to evaluate the algorithms in the user study. Red paths represent the mesh-based method, orange paths represent the image-based method, and blue paths represent the human-drawn paths.
    }
    \vspace{-0.3cm}
\end{figure*}

\section{Path Planning Evaluation}

To evaluate the path planning module's design choice of following curves of the hair as an appropriate combing strategy, we propose a baseline alternative, called the mesh-based method. In this approach, the strategy for combing is instead based on selecting paths that follow the shortest path to the bottom of the hair from an initial starting position.

\subsection{Mesh-Based Method}
To create paths that move directly downward, we plan directly on a mesh created from the point cloud of the hair, taken by the Kinect camera. Next we segment the image based on color and depth to find candidate point clouds of the head. A human operator selects the subset of point clouds that correspond to the user's hair. The union of these point clouds represent the entirety of the hair structure to be brushed. The resulting point cloud is then converted to a mesh by using a greedy surface triangulation algorithm \cite{Marton09ICRA} from the PointCloud Library \cite{Rusu_ICRA2011_PCL}.
Once the mesh is created, paths are formed by sampling mesh vertices of the hair mesh model as starting points. The mesh is a graph with nodes corresponding to mesh vertices and edges corresponding to the edges of the triangles that define the mesh. We define the goal as any point in the bottom portion of the hair mesh. We then formulate path planning as finding a low-cost path from the start vertex to one of the vertices in the goal set. We find these paths using the A* algorithm \cite{russell2002artificial} with Euclidean distance as the distance metric and the straight-line distance to the bottom of the mesh as the heuristic.


To generate a variety of paths, the algorithm is run several times using a variety of sampled start states. Examples of paths generated by this algorithm and how it compared to the image-based algorithm is shown in Figure \ref{fig:meshPaths}. The key difference between these algorithms is that the mesh-based algorithm combs in a downward manner, whereas the image-based algorithm is able to additionally brush sideways paths.

\subsection{Study Setup}
To evaluate the strengths and weaknesses of the these approaches, we measured user perception of the methods through an on-line study using Amazon Mechanical Turk. A variety of hairstyles were selected from the USC hair salon database \cite{hu2015single}. Ten short, ten medium, and ten long hairstyles were selected at random, and rendered in one of five hair colors (black, brown, auburn, blond(e), and grey) from one of three camera angles.  

Each MTurk participant saw 10 of the 30 hairstyles at random, and was shown three types of paths: the mesh-based method, the image-based method, and a set of human-drawn paths as a high-quality reference point. Human-drawn paths were generated by an independent researcher not affiliated with the project.
For each algorithm, participants were shown three paths randomly selected from a set of solutions at different starting points to provide a holistic view of the types of paths the algorithms generated. For each algorithm, participants rated the generated paths on: 1) whether the paths \textit{are able} to brush hair (referred to as \textbf{completeness}), and 2) whether the paths \textit{effectively} brush the hair (referred to as \textbf{effectiveness}). 


\subsection{Hypotheses}
\noindent
\textbf{H1}: \textit{Participants will prefer paths that vary in direction, and thus rate the image-based method as generating paths that are more complete than the mesh-based method.} 

\noindent
\textbf{H2}: \textit{Participants will prefer paths that follow the flow of the hair, and thus rate the image-based method as generating paths that are more effective than the mesh-based method.}\\ 
\vspace{-0.2cm}
\begin{figure}[t]
  \centering
  \includegraphics[width=\linewidth]{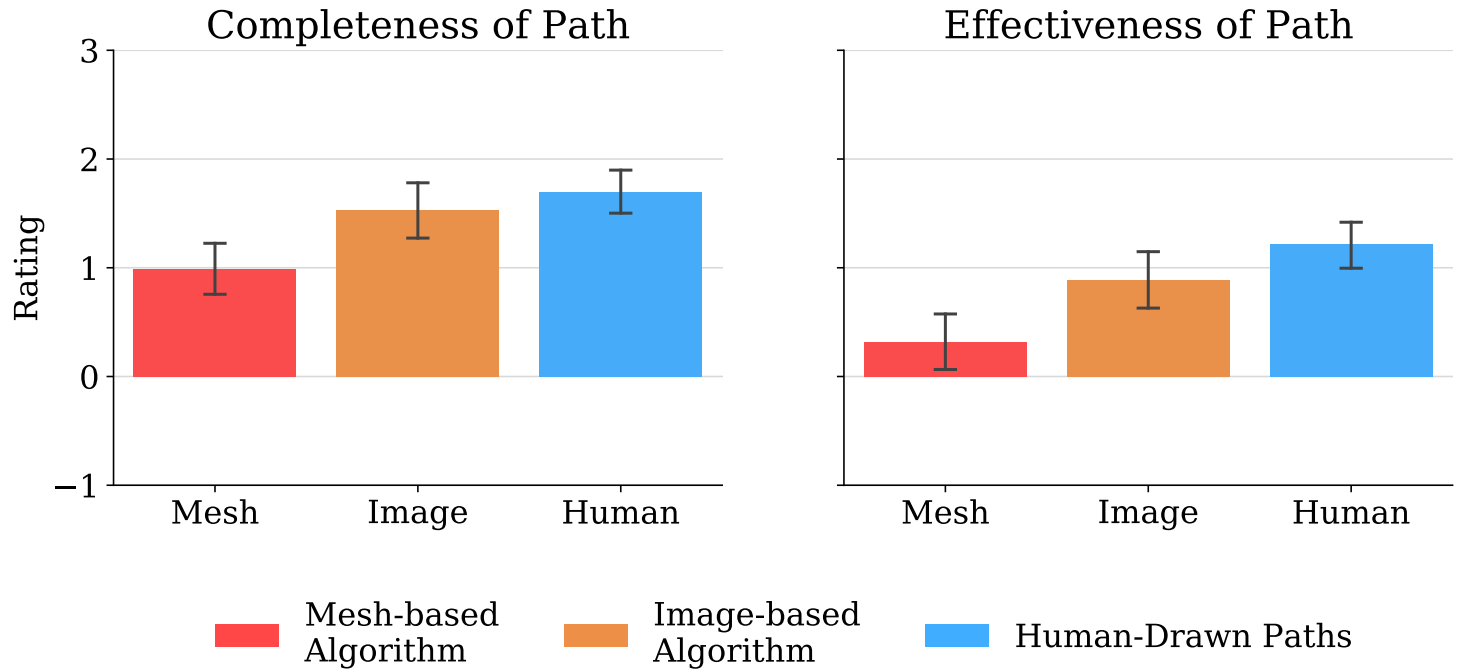}
  \caption{Means of the three different algorithms. Error bars represent 95\% Confidence Intervals of mean ratings. All differences are significant.}
   \label{fig:Evaluations}
   \vspace{-.5cm}
\end{figure}

\begin{table*}[t]
  \begin{centering}
  \begin{tabular}{ |c c c| }
\hline
 Theme & Positive Example & Negative Example \\ 
 \hline
 Effectiveness & ``...the hair [that] the robot worked on looked much better..." & ``...there is no effect on the hair that the robot is combing." \\  
 Force & ``It did seem gentle, so that is good." & ``...it looked like it was pushing rather rough at the hair." \\
 Movement & ``This robot appeared to have finesse and sensitive movements." & ``...there is no grace or nuance to the motion." \\
 \hline
\end{tabular}
  \caption{Examples of positive and negative qualitative responses from participants for each theme.}
  \label{tab:responses}
  \end{centering}
  \vspace{-0.3cm}
\end{table*}

\begin{figure*}[t]
  \centering
  
  \begin{subfigure}[c]{0.765\textwidth}
      \includegraphics[width=\linewidth]{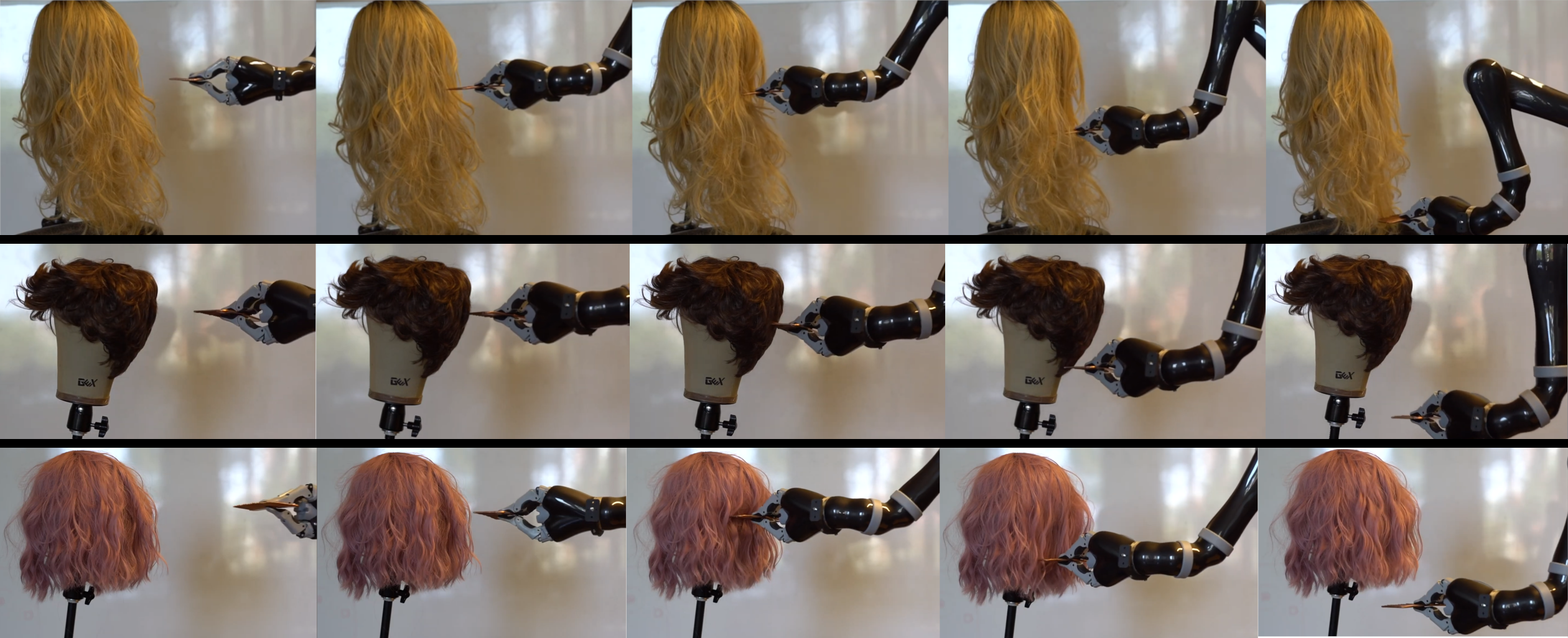}
    \end{subfigure}\hfil 
    \begin{subfigure}[c]{0.22\textwidth}
      \includegraphics[width=\linewidth]{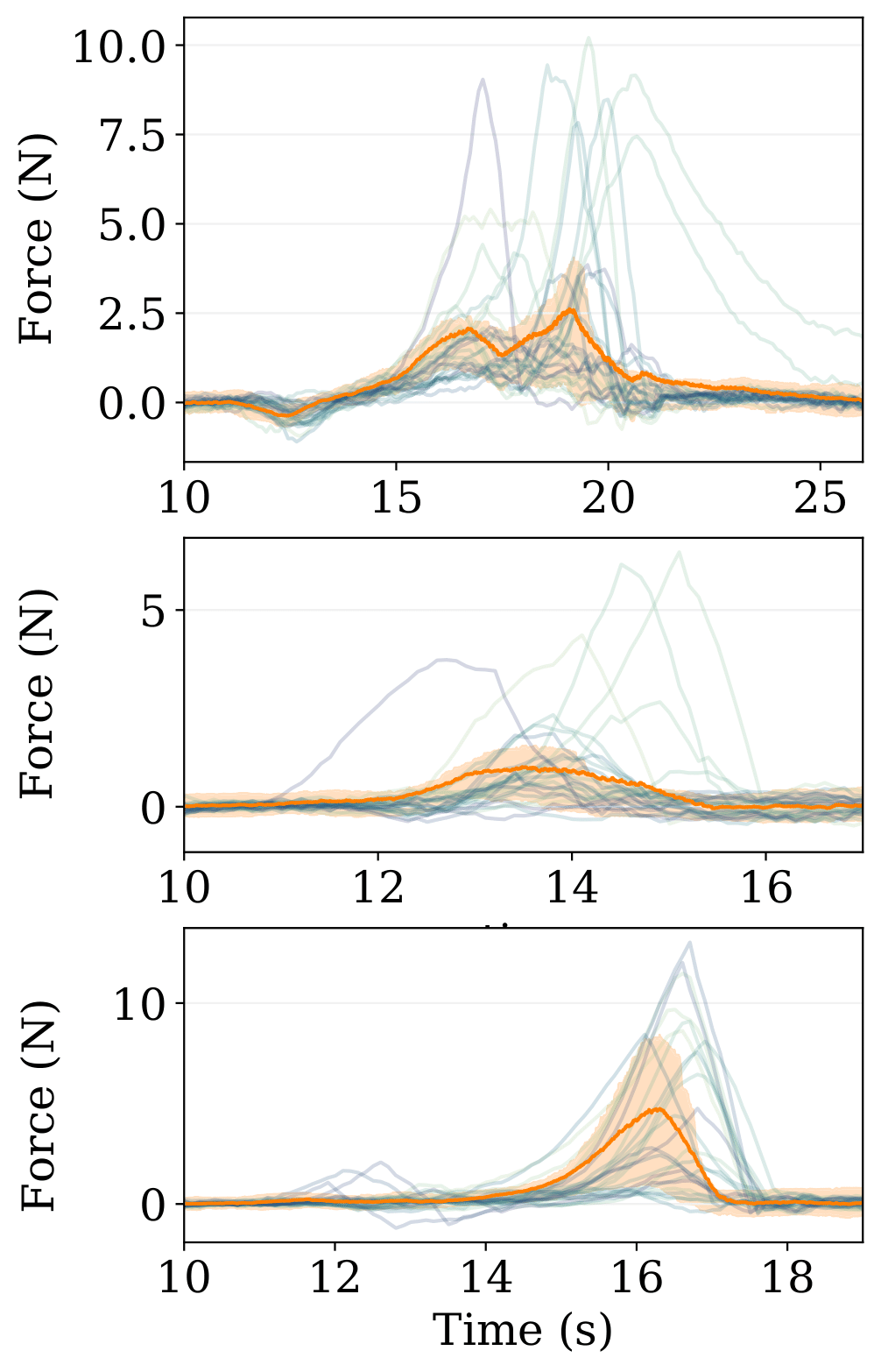}
    \end{subfigure}\hfil 

  \caption{Sample frames of the video shown to the participants and forces measured by the arm for different strokes of each hairstyle. Orange lines represent mean force values across 25 strokes with the orange region illustrating the first and third quartiles. Blue lines represent individual stroke force values. All force readings were measured at 10Hz and were post-processed with a sliding average of 9 timesteps for visualization.}
  \label{fig:studyFrame}
   \vspace{-0.5cm}
\end{figure*}
\subsection{Results}
The study received 48 respondents (27 women, 21 men; median age range 25-34). Responses were measured on a scale of -3 to 3, with 0 indicating neutral responses. A repeated-measures ANOVA test found significant differences for the main effect of algorithm on reported \textbf{completeness} of planned path, with F(2,94) = 24.12, $p<.001$. Pairwise t-tests were used for post-hoc analysis, revealing that all pairwise comparisons were significant after Bonferroni correction. The mesh-based method of planning paths (M=.99) received significantly lower ratings than the image-based method (M=1.53), $p<.001$, $\eta^2=.09$ which was in turn rated as less complete than the human-drawn paths (M=1.70), $p=.01$, $\eta^2=.01$. The difference between the mesh-based method and the human-drawn paths was also significant, $p<.001$, $\eta^2=.17$. This supports \textbf{H1}.

We performed the same analysis for \textbf{effectiveness} ratings. A repeated-measures ANOVA test shows significant differences by algorithm type, with F(2,94) = 21.69, $p<.001$. Pairwise t-tests show significance between all comparisons with $p<.001$, where the mesh-based method (M=.32) was lower than the image-based method (M=.89), $\eta^2=.08$, and the human-drawn paths (M=1.22) were rated the highest, $\eta^2=.04$. This supports \textbf{H2}.

Qualitative notes from participants in the study revealed that they found the human paths to be rated the most highly because they tended to ``follow the hair flow" but were also ``long and straight". The main concern with the image-based method was the it followed the path of the hair too closely. This is evident for the hairstyles that have waves, where combing in a wavy path may be inefficient. Because the hair can deform as the comb passes through it, moving along the waves is less direct than moving straight. However, moving along the waves is still preferable to moving directly downward without following the general direction of the hair, as in the mesh-based method.


\section{Trajectory Execution}

\subsection{On-line User Study}
To evaluate the performance of the physical robot implementation of the system, we recorded three videos of the robot combing three different wigs.\footnote{We are unable to perform in-person studies due to the ongoing pandemic.} The wigs varied in color and hairstyle and were representative of the hairstyles in the first study, consisting of a short brown wig, a medium-length pink wig, and a long blonde wig. The wigs were teased so that the hair was in a state that would warrant using a hair-combing device. Each video showed the wig and robot arm from the side so that study participants could see both the end effector and wig at all times, shown in Fig. \ref{fig:studyFrame}. We used a plain white background for all videos. A human operator selected the start points for the robot system off-screen. The participants watched the wig being combed by the robot, where the first and final strokes were filmed in real time and the middle strokes were filmed at 20x speed to show the progress that the robot makes over the course of the hair-combing interaction to reduce fatigue effects. Participants were then asked to provide qualitative feedback on the effectiveness of the combing as well as their willingness to use the system.

\subsection{Qualitative Analysis}
The study received 30 responses (17 men, 13 women; median age range 25-34). We conducted a qualitative analysis of the open-ended survey responses utilizing a grounded-theory approach \cite{heath2004developing} where we categorized responses by theme in a data-driven manner. 
The qualitative analysis revealed three main themes in participants' evaluations of the system. The first theme was related to the effectiveness of the task of hair combing. The second theme was the perception of the amount of force that was used throughout the interaction.  The third theme was related to how the robot moved while combing hair. Representative positive and negative examples of each theme are shown in Table~\ref{tab:responses}.

\subsubsection{Effectiveness}
The theme of effectiveness had several subcomponents. The top three were: final alignment of hair, depth of combing, and time it takes the robot to comb.

The final alignment of the hair was mentioned in 31 of the 90 responses; 25 of these responses indicated that respondents thought that the hair ended up ``...straighter than before it started combing" or some variant highlighting that the hair looked neater after the combing interaction. This subcomponent received the highest number of mentions across all categories, indicating that the robot is functionally capable of combing hair. 

Related to depth of combing, the majority of negative responses came from the short brown hairstyle due to the fact that the hair lies very close to the scalp. 19 responses mentioned that the robot appears to comb the hair on the outermost layer, but 4 of the responses indicate that the robot also penetrates the deeper layers of the hair. All 4 positive responses were for the medium and long hairstyles. 

The subcomponent of timing was noted in 24 responses. 19 of these responses indicated that the robot took a long time to move when combing, indicating that they were willing to have a robot that has higher velocity limits for this task. Interestingly, 4 responses positively described the slow movement of the robot as cautious and gentle.

\subsubsection{Use of Force}
The theme of force was concerned with the amount of force the robot used during the combing interaction. The most common comment was that the robot appeared to use too much force when combing the wig, present in 15 of the responses. This was either stated generally, or quantified as the amount of movement that the mannequin head made during the combing execution. This brought up concerns for a human user's safety when using such a system in 6 of the participants' responses. Conversely, the amount of force used was perceived as gentle and appropriate by 6 of the responses, who mentioned that the robot appeared to move carefully. Participants responded that the robot did not appear to have an understanding of the sensation of touch, which worried them as the robot may not realized how hard it is pulling on knots or tangles. Physically, the maximum force exerted by the system for any hairstyle did not exceed 15N, with the majority of strokes exerting peaks of less than 5N for each hair style as measured by joint efforts of the arm (see Fig. \ref{fig:studyFrame}). These values are similar in magnitude to sensitive physical interaction tasks such as shaving (7N peak magnitude) and wiping cheeks (14N peak magnitude) \cite{hawkins2012informing}.

\subsubsection{Characteristics of Movement}
The responses that regarded the characteristics of movement mentioned both auditory and visual components. Participants mentioned the lack of sound during the movement as uncomfortable. Visually, some participants described the static orientation of the hand during the approach as lacking nuance, whereas others indicated that the robot following the curves and shape of the hair as having finesse and being smooth. This indicates a need to adapt motion to the specific user of the system.

\section{Discussion and Conclusion}




Results from the qualitative survey indicate that participants found the proposed system to be effective in hair combing, but they raised interesting points to improve the user's experience during the interaction.

A key limitation of this study is that it was conducted through online videos and not in person, limiting participants to only visual and auditory modalities of perception. This allowed them to accurately make judgements of the sound that the robot makes while moving, but prevented them from directly experiencing tactile information such as forces used by the system. While the robot in the videos implemented position control, our system provides only the set of goal positions and their associated times. Implementing hybrid position-force controllers \cite{raibert1981hybrid} would allow the system to move between these same goal positions while simultaneously enforcing force constraints, allowing users to specify the force that they are comfortable with. Being controller-agnostic additionally allows the force to be estimated through different kinds of sensors depending on the robot (e.g., current sensors, end effector force sensors, etc.). Further work should critically evaluate controller design and user acceptance in a real-world setting.  

Participants indicated concern that the silence of the interaction was unsettling, reflecting findings of robots in social settings \cite{hawkins2012informing, chen2013robots, cha2015perceived}, but contrary to findings in other impersonal and intimate interaction settings \cite{chen2011touched,trovato2018sound}. Future work should explore the implementation of pleasant auditory elements that allow the anticipation of the robot's motion. 


Some participants additionally indicated that the qualitative aspects of the robot's movement lacked nuance. Previous works have linked user perception of robot movement to the orientation and speed of the robot \cite{knight2016expressive,rond2019improv}, warranting future work in modulating the end effector's orientation and speed during its approach. For use in intended assistive services, the robot's physical appearance can also be improved with best embodiment design practices, as discussed in Deng et al. \cite{deng2019embodiment}, e.g., using robotic arms with a traditionally less mechanical appearances for these social tasks.




 While our algorithm performs well on the hairstyles selected from the database, the evaluation does not cover all possible hairstyles, such as very curly hair or sparse hair which are realistic to encounter in naturalistic settings. However, it is important to consider that not all hairstyles should be dry-combed daily; for example, combing very curly hair may damage coils, and certain styles such as buzz cuts or braided hair would not benefit from utilizing this system. Nevertheless, validating this approach on other relevant hairstyles is an important direction for future work. 


Overall, this work brings about a better understanding of how general-purpose robotic arms can effectively comb hair by adapting to different hairstyles. We presented a modular system for robotic hair combing using a minimal setup that can work in conjunction with other assistive applications. We found that users' experience is related to the robot's movement, sound, and perceived applied force, thus opening up a range of human-robot interaction research questions. 




\bibliographystyle{ieeetr}
\bibliography{references}

\begin{thebibliography}{10}

\bibitem{nhis_2019}
NHIS, ``Table a-10. difficulties in physical functioning among adults aged 18
  and over, by selected characteristics: United states, 2018,'' Nov 2019.

\bibitem{feinberg2011valuing}
L.~Feinberg, S.~C. Reinhard, A.~Houser, R.~Choula, {\em et~al.}, ``Valuing the
  invaluable: 2011 update, the growing contributions and costs of family
  caregiving,'' {\em Washington, DC: AARP Public Policy Institute}, vol.~32,
  p.~2011, 2011.

\bibitem{pinquart2003differences}
M.~Pinquart and S.~S{\"o}rensen, ``Differences between caregivers and
  noncaregivers in psychological health and physical health: a
  meta-analysis.,'' {\em Psychology and aging}, vol.~18, no.~2, p.~250, 2003.

\bibitem{lin2011does}
I.-F. Lin and H.-S. Wu, ``Does informal care attenuate the cycle of adl/iadl
  disability and depressive symptoms in late life?,'' {\em Journals of
  Gerontology Series B: Psychological Sciences and Social Sciences}, vol.~66,
  no.~5, pp.~585--594, 2011.

\bibitem{bhattacharjee2020more}
T.~Bhattacharjee, E.~K. Gordon, R.~Scalise, M.~E. Cabrera, A.~Caspi, M.~Cakmak,
  and S.~S. Srinivasa, ``Is more autonomy always better? exploring preferences
  of users with mobility impairments in robot-assisted feeding,'' in {\em
  Proceedings of the 2020 ACM/IEEE International Conference on Human-Robot
  Interaction}, pp.~181--190, 2020.

\bibitem{goldau2019autonomous}
F.~F. Goldau, T.~K. Shastha, M.~Kyrarini, and A.~Gr{\"a}ser, ``Autonomous
  multi-sensory robotic assistant for a drinking task,'' in {\em 2019 IEEE 16th
  International Conference on Rehabilitation Robotics (ICORR)}, pp.~210--216,
  IEEE, 2019.

\bibitem{jain2010assistive}
A.~Jain and C.~C. Kemp, ``El-e: an assistive mobile manipulator that
  autonomously fetches objects from flat surfaces,'' {\em Autonomous Robots},
  vol.~28, no.~1, p.~45, 2010.

\bibitem{hawkins2014assistive}
K.~P. Hawkins, P.~M. Grice, T.~L. Chen, C.-H. King, and C.~C. Kemp, ``Assistive
  mobile manipulation for self-care tasks around the head,'' in {\em 2014 IEEE
  Symposium on computational intelligence in robotic rehabilitation and
  assistive technologies (CIR2AT)}, pp.~16--25, IEEE, 2014.

\bibitem{king2010towards}
C.-H. King, T.~L. Chen, A.~Jain, and C.~C. Kemp, ``Towards an assistive robot
  that autonomously performs bed baths for patient hygiene,'' in {\em 2010
  IEEE/RSJ International Conference on Intelligent Robots and Systems},
  pp.~319--324, IEEE, 2010.

\bibitem{erickson2018deep}
Z.~Erickson, H.~M. Clever, G.~Turk, C.~K. Liu, and C.~C. Kemp, ``Deep haptic
  model predictive control for robot-assisted dressing,'' in {\em 2018 IEEE
  international conference on robotics and automation (ICRA)}, pp.~1--8, IEEE,
  2018.

\bibitem{kapusta2019personalized}
A.~Kapusta, Z.~Erickson, H.~M. Clever, W.~Yu, C.~K. Liu, G.~Turk, and C.~C.
  Kemp, ``Personalized collaborative plans for robot-assisted dressing via
  optimization and simulation,'' {\em Autonomous Robots}, vol.~43, no.~8,
  pp.~2183--2207, 2019.

\bibitem{brose2010role}
S.~W. Brose, D.~J. Weber, B.~A. Salatin, G.~G. Grindle, H.~Wang, J.~J. Vazquez,
  and R.~A. Cooper, ``The role of assistive robotics in the lives of persons
  with disability,'' {\em American Journal of Physical Medicine \&
  Rehabilitation}, vol.~89, no.~6, pp.~509--521, 2010.

\bibitem{candeias2018vision}
A.~Candeias, T.~Rhodes, M.~Marques, M.~Veloso, {\em et~al.}, ``Vision augmented
  robot feeding,'' in {\em PROCEEDINGS of the European Conference on Computer
  Vision (ECCV)}, pp.~0--0, 2018.

\bibitem{jardon2012functional}
A.~Jard{\'o}n, C.~A. Monje, and C.~Balaguer, ``Functional evaluation of asibot:
  A new approach on portable robotic system for disabled people,'' {\em Applied
  Bionics and Biomechanics}, vol.~9, no.~1, pp.~85--97, 2012.

\bibitem{alqasemi2005wheelchair}
R.~M. Alqasemi, E.~J. McCaffrey, K.~D. Edwards, and R.~V. Dubey,
  ``Wheelchair-mounted robotic arms: Analysis, evaluation and development,'' in
  {\em Proceedings, 2005 IEEE/ASME International Conference on Advanced
  Intelligent Mechatronics.}, pp.~1164--1169, IEEE, 2005.

\bibitem{schrock2009design}
P.~Schrock, F.~Farelo, R.~Alqasemi, and R.~Dubey, ``Design, simulation and
  testing of a new modular wheelchair mounted robotic arm to perform activities
  of daily living,'' in {\em 2009 IEEE International Conference on
  Rehabilitation Robotics}, pp.~518--523, IEEE, 2009.

\bibitem{gallenberger2019transfer}
D.~Gallenberger, T.~Bhattacharjee, Y.~Kim, and S.~S. Srinivasa, ``Transfer
  depends on acquisition: Analyzing manipulation strategies for robotic
  feeding,'' in {\em 2019 14th ACM/IEEE International Conference on Human-Robot
  Interaction (HRI)}, pp.~267--276, IEEE, 2019.

\bibitem{park2016towards}
D.~Park, Y.~K. Kim, Z.~M. Erickson, and C.~C. Kemp, ``Towards assistive feeding
  with a general-purpose mobile manipulator,'' {\em arXiv preprint
  arXiv:1605.07996}, 2016.

\bibitem{rhodes2018robot}
T.~Rhodes and M.~Veloso, ``Robot-driven trajectory improvement for feeding
  tasks,'' in {\em 2018 IEEE/RSJ International Conference on Intelligent Robots
  and Systems (IROS)}, pp.~2991--2996, IEEE, 2018.

\bibitem{kinova}
Kinova, ``Assistive solutions.''

\bibitem{performance}
P.~Health, ``Obi robotic feeder.''

\bibitem{assistive}
A.~Innovations, ``Assistive innovations - iarm: Robotic arm for humans,
  mountable on powered wheelchair.''

\bibitem{herlant2016assistive}
L.~V. Herlant, R.~M. Holladay, and S.~S. Srinivasa, ``Assistive teleoperation
  of robot arms via automatic time-optimal mode switching,'' in {\em 2016 11th
  ACM/IEEE International Conference on Human-Robot Interaction (HRI)},
  pp.~35--42, IEEE, 2016.

\bibitem{schukat2016unintended}
M.~Schukat, D.~McCaldin, K.~Wang, G.~Schreier, N.~H. Lovell, M.~Marschollek,
  and S.~J. Redmond, ``Unintended consequences of wearable sensor use in
  healthcare: Contribution of the imia wearable sensors in healthcare wg,''
  {\em Yearbook of medical informatics}, no.~1, p.~73, 2016.

\bibitem{ward2007survey}
K.~Ward, F.~Bertails, T.-Y. Kim, S.~R. Marschner, M.-P. Cani, and M.~C. Lin,
  ``A survey on hair modeling: Styling, simulation, and rendering,'' {\em IEEE
  transactions on visualization and computer graphics}, vol.~13, no.~2,
  pp.~213--234, 2007.

\bibitem{saito20183d}
S.~Saito, L.~Hu, C.~Ma, H.~Ibayashi, L.~Luo, and H.~Li, ``3d hair synthesis
  using volumetric variational autoencoders,'' {\em ACM Transactions on
  Graphics (TOG)}, vol.~37, no.~6, pp.~1--12, 2018.

\bibitem{zhang2019hair}
M.~Zhang and Y.~Zheng, ``Hair-gan: Recovering 3d hair structure from a single
  image using generative adversarial networks,'' {\em Visual Informatics},
  vol.~3, no.~2, pp.~102--112, 2019.

\bibitem{hu2015single}
L.~Hu, C.~Ma, L.~Luo, and H.~Li, ``Single-view hair modeling using a hairstyle
  database,'' {\em ACM Transactions on Graphics (ToG)}, vol.~34, no.~4,
  pp.~1--9, 2015.

\bibitem{chai2012single}
M.~Chai, L.~Wang, Y.~Weng, Y.~Yu, B.~Guo, and K.~Zhou, ``Single-view hair
  modeling for portrait manipulation,'' {\em ACM Transactions on Graphics
  (TOG)}, vol.~31, no.~4, pp.~1--8, 2012.

\bibitem{maheu2011evaluation}
V.~Maheu, P.~S. Archambault, J.~Frappier, and F.~Routhier, ``Evaluation of the
  jaco robotic arm: Clinico-economic study for powered wheelchair users with
  upper-extremity disabilities,'' in {\em 2011 IEEE International Conference on
  Rehabilitation Robotics}, pp.~1--5, IEEE, 2011.

\bibitem{haralick1985image}
R.~M. Haralick and L.~G. Shapiro, ``Image segmentation techniques,'' {\em
  Computer vision, graphics, and image processing}, vol.~29, no.~1,
  pp.~100--132, 1985.

\bibitem{ghosh2019understanding}
S.~Ghosh, N.~Das, I.~Das, and U.~Maulik, ``Understanding deep learning
  techniques for image segmentation,'' {\em ACM Computing Surveys (CSUR)},
  vol.~52, no.~4, pp.~1--35, 2019.

\bibitem{iandola2016squeezenet}
F.~N. Iandola, S.~Han, M.~W. Moskewicz, K.~Ashraf, W.~J. Dally, and K.~Keutzer,
  ``Squeezenet: Alexnet-level accuracy with 50x fewer parameters and< 0.5 mb
  model size,'' {\em arXiv preprint arXiv:1602.07360}, 2016.

\bibitem{umar2018hair}
U.~R. Muhammad, M.~Svanera, R.~Leonardi, and S.~Benini, ``Hair detection,
  segmentation, and hairstyle classification in the wild,'' {\em Image and
  Vision Computing}, 2018.

\bibitem{weickert2003coherence}
J.~Weickert, ``Coherence-enhancing shock filters,'' in {\em Joint Pattern
  Recognition Symposium}, pp.~1--8, Springer, 2003.

\bibitem{yang1996structure}
G.-Z. Yang, P.~Burger, D.~N. Firmin, and S.~Underwood, ``Structure adaptive
  anisotropic image filtering,'' {\em Image and Vision Computing}, vol.~14,
  no.~2, pp.~135--145, 1996.

\bibitem{Marton09ICRA}
Z.~C. Marton, R.~B. Rusu, and M.~Beetz, ``{On Fast Surface Reconstruction
  Methods for Large and Noisy Datasets},'' in {\em Proceedings of the IEEE
  International Conference on Robotics and Automation (ICRA)}, (Kobe, Japan),
  May 12-17 2009.

\bibitem{Rusu_ICRA2011_PCL}
R.~B. Rusu and S.~Cousins, ``{3D is here: Point Cloud Library (PCL)},'' in {\em
  {IEEE International Conference on Robotics and Automation (ICRA)}},
  (Shanghai, China), May 9-13 2011.

\bibitem{russell2002artificial}
S.~Russell and P.~Norvig, ``Artificial intelligence: a modern approach,'' 2002.

\bibitem{heath2004developing}
H.~Heath and S.~Cowley, ``Developing a grounded theory approach: a comparison
  of glaser and strauss,'' {\em International journal of nursing studies},
  vol.~41, no.~2, pp.~141--150, 2004.

\bibitem{hawkins2012informing}
K.~P. Hawkins, C.-H. King, T.~L. Chen, and C.~C. Kemp, ``Informing assistive
  robots with models of contact forces from able-bodied face wiping and
  shaving,'' in {\em 2012 IEEE RO-MAN: The 21st IEEE International Symposium on
  Robot and Human Interactive Communication}, pp.~251--258, IEEE, 2012.

\bibitem{raibert1981hybrid}
M.~H. Raibert and J.~J. Craig, ``Hybrid position/force control of
  manipulators,'' 1981.

\bibitem{chen2013robots}
T.~L. Chen, M.~Ciocarlie, S.~Cousins, P.~M. Grice, K.~Hawkins, K.~Hsiao, C.~C.
  Kemp, C.-H. King, D.~A. Lazewatsky, A.~E. Leeper, {\em et~al.}, ``Robots for
  humanity: using assistive robotics to empower people with disabilities,''
  {\em IEEE Robotics \& Automation Magazine}, vol.~20, no.~1, pp.~30--39, 2013.

\bibitem{cha2015perceived}
E.~Cha, A.~D. Dragan, and S.~S. Srinivasa, ``Perceived robot capability,'' in
  {\em 2015 24th IEEE International Symposium on Robot and Human Interactive
  Communication (RO-MAN)}, pp.~541--548, IEEE, 2015.

\bibitem{chen2011touched}
T.~L. Chen, C.-H. King, A.~L. Thomaz, and C.~C. Kemp, ``Touched by a robot: An
  investigation of subjective responses to robot-initiated touch,'' in {\em
  2011 6th ACM/IEEE International Conference on Human-Robot Interaction (HRI)},
  pp.~457--464, IEEE, 2011.

\bibitem{trovato2018sound}
G.~Trovato, R.~Paredes, J.~Balvin, F.~Cuellar, N.~B. Thomsen, S.~Bech, and
  Z.-H. Tan, ``The sound or silence: investigating the influence of robot noise
  on proxemics,'' in {\em 2018 27th IEEE International Symposium on Robot and
  Human Interactive Communication (RO-MAN)}, pp.~713--718, IEEE, 2018.

\bibitem{knight2016expressive}
H.~Knight, R.~Thielstrom, and R.~Simmons, ``Expressive path shape (swagger):
  Simple features that illustrate a robot's attitude toward its goal in real
  time,'' in {\em 2016 IEEE/RSJ International Conference on Intelligent Robots
  and Systems (IROS)}, pp.~1475--1482, IEEE, 2016.

\bibitem{rond2019improv}
J.~Rond, A.~Sanchez, J.~Berger, and H.~Knight, ``Improv with robots:
  Creativity, inspiration, co-performance,'' in {\em 2019 28th IEEE
  International Conference on Robot and Human Interactive Communication
  (RO-MAN)}, pp.~1--8, IEEE, 2019.

\bibitem{deng2019embodiment}
E.~Deng, B.~Mutlu, and M.~Mataric, ``Embodiment in socially interactive
  robots,'' {\em arXiv preprint arXiv:1912.00312}, 2019.

\end{thebibliography}

\end{document}